\documentclass[sigconf]{acmart}

\usepackage{graphicx}
\usepackage{algpseudocode}
\usepackage[noend,ruled,linesnumbered]{algorithm2e}
\usepackage{subfigure}
\usepackage{array}
\usepackage{setspace}
\usepackage{float}
\usepackage{url}
\usepackage{balance}
\usepackage{amsfonts}
\usepackage{amsmath}
\usepackage{rotating}
\usepackage{multirow}
\usepackage{color}
\usepackage{bm}
\usepackage{enumitem}
\usepackage{pifont}
\usepackage{graphbox}

\definecolor{myblue}{rgb}{0.2, 0.2, 0.9}

\newcommand{\model}{MICoL}

\newcommand{\bme}{{\bm e}}
\newcommand{\bmw}{{\bm w}}
\newcommand{\bmy}{{\bm y}}

\newcommand\blfootnote[1]{%
  \begingroup
  \renewcommand\thefootnote{}\footnote{#1}%
  \addtocounter{footnote}{-1}%
  \endgroup
}

\AtBeginDocument{%
  \providecommand\BibTeX{{%
    \normalfont B\kern-0.5em{\scshape i\kern-0.25em b}\kern-0.8em\TeX}}}

\copyrightyear{2022}
\acmYear{2022}
\setcopyright{acmcopyright}\acmConference[WWW '22]{Proceedings of the ACM Web Conference 2022}{April 25--29, 2022}{Virtual Event, Lyon, France}
\acmBooktitle{Proceedings of the ACM Web Conference 2022 (WWW '22), April 25--29, 2022, Virtual Event, Lyon, France}
\acmPrice{15.00}
\acmDOI{10.1145/3485447.3512174}
\acmISBN{978-1-4503-9096-5/22/04}

\begin{document}
\title{Metadata-Induced Contrastive Learning for \\ Zero-Shot Multi-Label Text Classification}

\author{Yu Zhang$^1$, Zhihong Shen$^2$, Chieh-Han Wu$^2$, Boya Xie$^2$, Junheng Hao$^3$, \\ Ye-Yi Wang$^2$, Kuansan Wang$^2$, Jiawei Han$^1$}
\affiliation{
\institution{$^1$University of Illinois at Urbana-Champaign\ \ \ \ \ $^2$Microsoft\ \ \ \ \ $^3$University of California, Los Angeles}
\country{$^1$\{yuz9, hanj\}@illinois.edu,\ \ \ \ \ $^2$\{zhihosh, chiewu, boxie, yeyiwang, kuansanw\}@microsoft.com,\ \ \ \ \ $^3$jhao@cs.ucla.edu}
}

\renewcommand{\shortauthors}{Zhang et al.}

\begin{abstract}
Large-scale multi-label text classification (LMTC) aims to associate a document with its relevant labels from a large candidate set. Most existing LMTC approaches rely on massive human-annotated training data, which are often costly to obtain and suffer from a long-tailed label distribution (i.e., many labels occur only a few times in the training set). In this paper, we study LMTC under the \textit{zero-shot} setting, which does not require any annotated documents with labels and only relies on label surface names and descriptions. To train a classifier that calculates the similarity score between a document and a label, we propose a novel metadata-induced contrastive learning (\textsc{\model}) method. Different from previous text-based contrastive learning techniques, \textsc{\model} exploits \textit{document metadata} (e.g., authors, venues, and references of research papers), which are widely available on the Web, to derive similar document--document pairs. Experimental results on two large-scale datasets show that: (1) \textsc{\model} significantly outperforms strong zero-shot text classification and contrastive learning baselines; (2) \textsc{\model} is on par with the state-of-the-art supervised metadata-aware LMTC method trained on 10K--200K labeled documents; and (3) \textsc{\model} tends to predict more infrequent labels than supervised methods, thus alleviates the deteriorated performance on long-tailed labels.
\blfootnote{$^*$Work performed while Yu and Junheng interned at Microsoft.}
\blfootnote{$^\dagger$The code and datasets are available at \textcolor{myblue}{\url{https://github.com/yuzhimanhua/MICoL}}.}
\end{abstract}



\begin{spacing}{0.97}

\maketitle

\section{Introduction}
Large-scale multi-label text classification (LMTC) \cite{chalkidis2019large} aims to find the most relevant labels to an input document given a large collection of candidate labels. As a fundamental task in text mining, LMTC has many Web-related applications such as product keyword recommendation on Amazon \cite{chang2020taming}, academic paper classification on Microsoft Academic and PubMed \cite{zhang2021match}, and article tagging on Wikipedia \cite{jain2016extreme}. 

Most previous attempts address LMTC in a supervised fashion \cite{agrawal2013multi,jain2016extreme,prabhu2018parabel,you2019attentionxml,jain2019slice,liu2017deep,yen2017ppdsparse,medini2019extreme,chang2020taming,zhang2021match,saini2021galaxc,mittal2021eclare}, where the proposed text classifiers are trained on a large set of human-annotated documents. While achieving inspiring performance, these approaches have three limitations. First, obtaining enough human-labeled training data is often expensive and time-consuming, especially when the label space is large. Second, the trained classifiers can only predict labels they have seen in the training set. When new categories (e.g., ``\texttt{COVID-19}'') emerge, the classifiers need to be re-trained. Third, the label distribution is often imbalanced in LMTC. Several labels (e.g., ``\texttt{World Wide Web}'') have numerous training samples, while many others (e.g., ``\texttt{Bipartite Ranking}'') occur only a few times. Related studies \cite{wei2018does,wei2021towards} have shown that supervised approaches tend to predict frequent labels and overlook long-tailed ones.

\begin{figure}
\centering
\subfigure[Label ``\texttt{Webgraph}'' from Microsoft Academic (\url{https://academic.microsoft.com/topic/2777569578/}).]{
\includegraphics[width=0.46\textwidth]{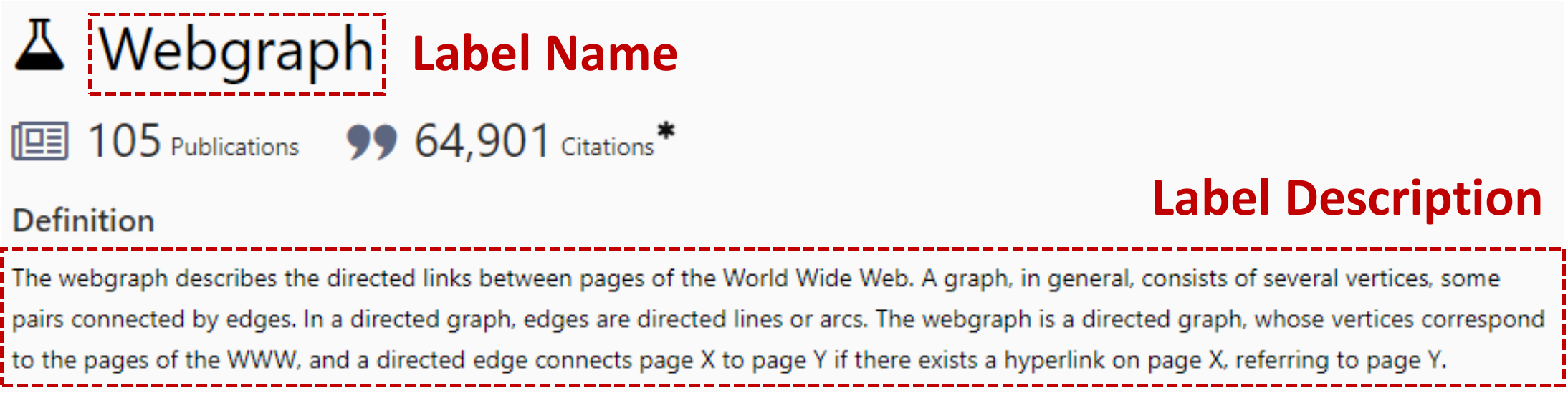}} \\
\vspace{-0.5em}
\subfigure[Label ``\texttt{Betacoronavirus}'' from PubMed (\url{https://meshb.nlm.nih.gov/record/ui?ui=D000073640}).]{
\includegraphics[width=0.46\textwidth]{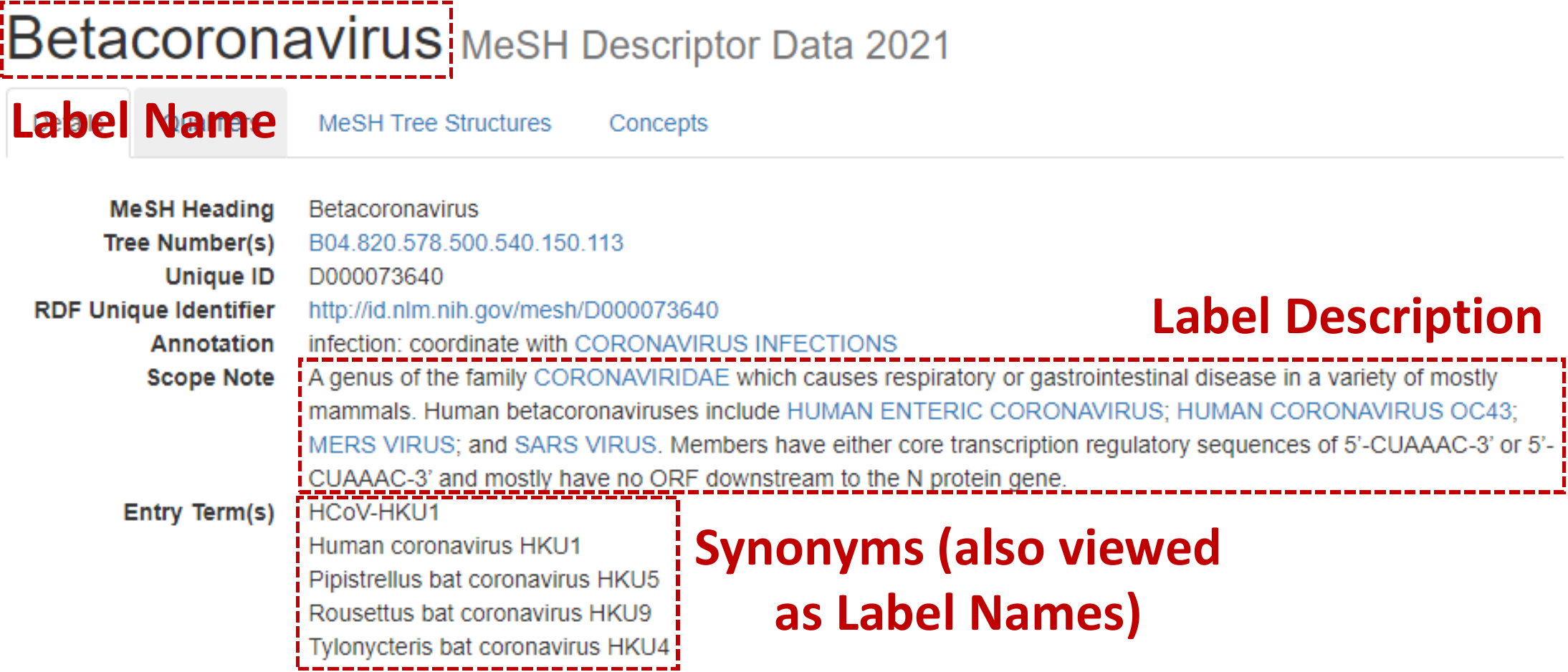}}
\vspace{-1em}
\caption{Two examples of labels with name(s) and description from Microsoft Academic \cite{wang2020microsoft} and PubMed \cite{lu2011pubmed}.} 
\vspace{-1em}
\label{fig:label}
\end{figure}

\begin{figure}
\centering
\subfigure[Training phase of \textsc{\model}.]{
\includegraphics[width=0.45\textwidth]{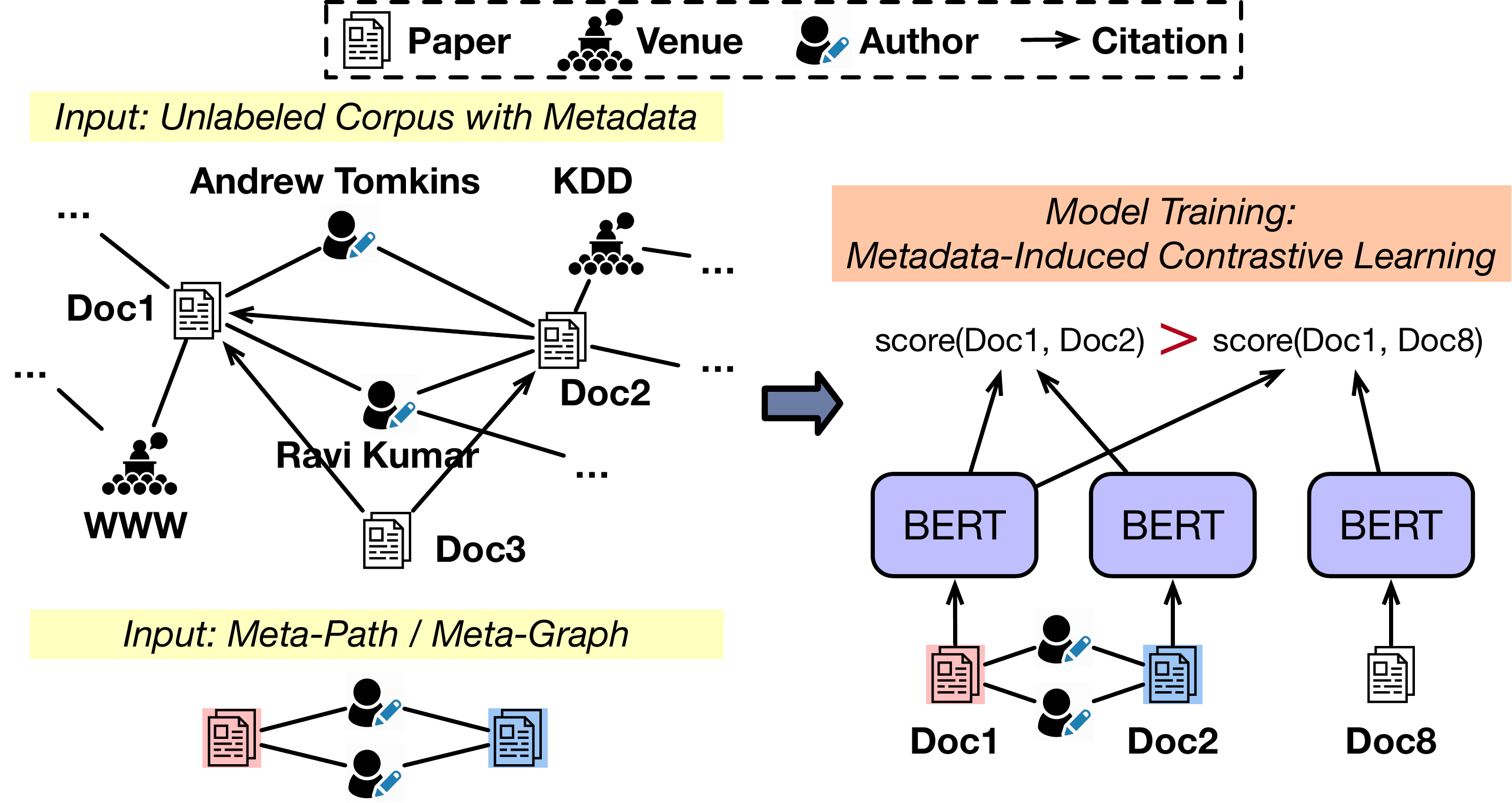}} \\
\vspace{-0.3em}
\subfigure[Inference phase of \textsc{\model}.]{
\includegraphics[width=0.47\textwidth]{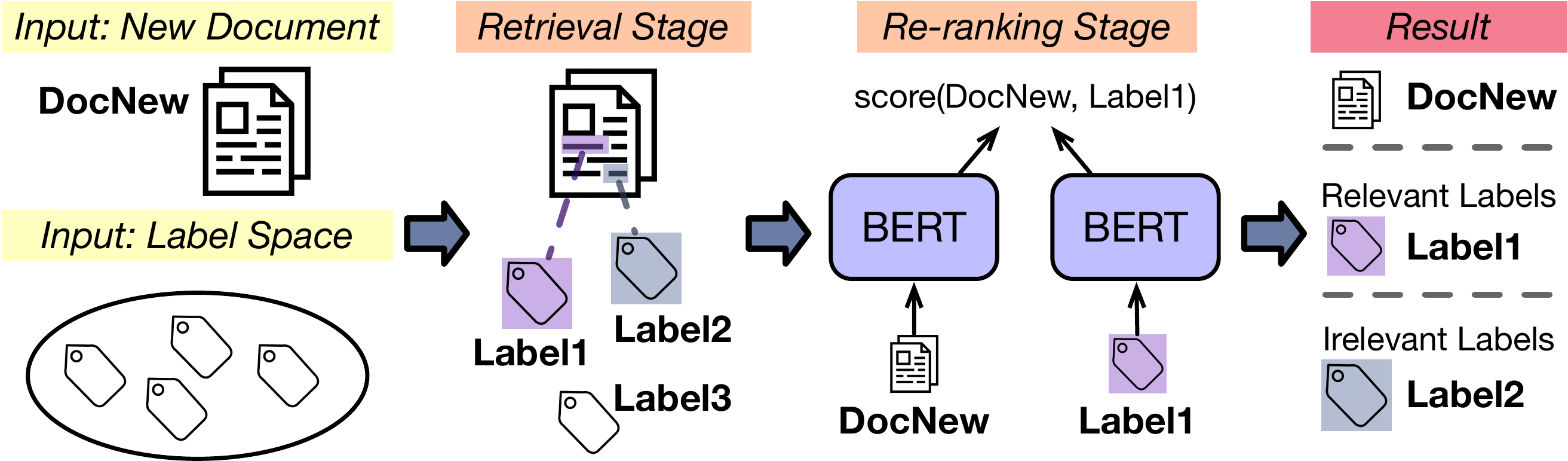}}
\vspace{-1em}
\caption{Overview of the proposed \textsc{\model} framework.} 
\vspace{-1em}
\label{fig:intro}
\end{figure}

Being aware of the annotation cost and frequent emergence of new labels, some studies \cite{chalkidis2019large,gupta2021generalized,nam2015predicting,rios2018few} focus on zero-shot LMTC. In their settings, annotated training documents are given for a set of \textit{seen} classes, and they are tasked to build a classifier to predict \textit{unseen} classes. However, as indicated in \cite{yin2019benchmarking}, we often face a more challenging scenario in real-world settings: all labels are \textit{unseen}; we do not have any training samples with labels. 
For example, on Microsoft Academic \cite{wang2020microsoft}, all labels are scientific concepts extracted from research publications on the Web \cite{shen2018web}, thus no annotated training data is available when the label space is created.
Motivated by such applications, in this paper, we study zero-shot LMTC with a completely new label space. The only signal we use to characterize each label is its surface name and description. 
Figure \ref{fig:label} shows two examples of label information from Microsoft Academic \cite{wang2020microsoft} and PubMed \cite{lu2011pubmed}.

Although relevant labels are not available for documents, there is another type of information widely available on the Web but less concerned in previous studies: \textit{document metadata}. We take scientific papers as an example: in addition to text information (e.g., title and abstract), a paper is also associated with various metadata fields such as its authors, venue, and references. As shown in Figure \ref{fig:intro}(a), a heterogeneous network \cite{sun2012mining} can be constructed to interconnect papers via their metadata information: papers, authors, and venues are nodes; authored-by, published-in, and cited-by are edges. Such metadata information could be strong label indicators of a paper. For example, in Figure \ref{fig:intro}(a), the venue node \textit{WWW} suggests \textit{Doc1}'s relevance to ``\texttt{World Wide Web}''. Moreover, metadata could also imply that two papers share common labels. For example, we know \textit{Doc1} and \textit{Doc2} may have similar research topics because they are co-authored by \textit{Andrew Tomkins} and \textit{Ravi Kumar} or they are co-cited by \textit{Doc3}. More generally, metadata also exist in Web content such as e-commerce reviews (e.g., reviewer and product information) \cite{zhang2021minimally}, social media posts (e.g., users and hashtags) \cite{zhang2017rate}, and code repositories (e.g., contributors) \cite{zhang2019higitclass}. Although metadata have been used in fully supervised \cite{zhang2021match} or single-label \cite{zhang2020minimally,mekala2020meta,zhang2021minimally,zhang2021motifclass} text classification, it is largely unexplored in zero-shot LMTC.

\vspace{1mm}

\noindent \textbf{Contributions.} In this paper, we propose a novel metadata-induced contrastive learning (\textsc{\model}) framework for zero-shot LMTC. To perform classification, the key module in our framework is to compute a similarity score between two text units (i.e., one document and one label name/description) so that we can produce a rank list of relevant labels for each document. Without annotated document--label pairs to train the similarity scorer, we leverage metadata to generate similar document--document pairs. Inspired by the idea of contrastive learning \cite{chen2020simple}, we train the scorer by pulling similar document closer while pushing dissimilar ones apart. For example, in Figure \ref{fig:intro}(a), we assume two papers sharing at least two authors are similar (this can be described by the notion of meta-paths \cite{sun2011pathsim} and meta-graphs \cite{zhang2018metagraph2vec}, which will be formally introduced in Section \ref{sec:meta}). The similarity scorer is trained to score (\textit{Doc1}, \textit{Doc2}) higher than (\textit{Doc1}, \textit{Doc8}), where \textit{Doc8} is a randomly sampled paper. In the inference phase, as shown in Figure \ref{fig:intro}(b), we first use a discrete retriever (e.g., BM25 \cite{robertson1994some}) to select a set of candidate labels from the large label space. Next, we utilize the trained scorer to re-rank candidate labels to obtain the final classification results. Note that label information is only used during inference, thus no re-training is required when new labels emerge.

We demonstrate the effectiveness of \textsc{\model} on two datasets \cite{zhang2021match} extracted from Microsoft Academic \cite{wang2020microsoft} and PubMed \cite{lu2011pubmed}, both with more than 15K labels. The results indicate that: (1) \textsc{\model} significantly outperforms strong zero-shot LMTC \cite{yin2019benchmarking} and contrastive learning \cite{wei2019eda,xie2020unsupervised,cohan2020specter} baselines. (2) When we use P@$k$ and NDCG@$k$ as evaluation metrics, \textsc{\model} is competitive with the state-of-the-art supervised metadata-aware LMTC algorithm \cite{zhang2021match} trained on 10K--50K labeled documents; (3) When it is evaluated by metrics promoting correct prediction on tail labels  \cite{jain2016extreme,wei2021towards}, \textsc{\model} is on par with the supervised method trained on 100K--200K labeled documents. This demonstrates that \textsc{\model} tends to predict more infrequent labels than supervised methods, thus alleviates the deteriorated performance on tail labels.

To summarize, this work makes the following contributions:
\begin{itemize}[leftmargin=*]
    \item We propose a zero-shot LMTC framework that utilizes document metadata. The framework does not require any labeled training data and only relies on label surface names and descriptions during inference.
    \item We propose a novel metadata-induced contrastive learning method. Different from previous contrastive learning approaches \cite{giorgi2021declutr,wu2020clear,luo2021unsupervised,gao2021simcse,yan2021consert} which manipulate text only, we exploit metadata information to produce contrastive training pairs.
    \item We conduct extensive experiments on two large-scale datasets to demonstrate the effectiveness of the proposed \textsc{\model} framework. 
\end{itemize}

\section{Preliminaries}
\subsection{Metadata, Meta-Path, and Meta-Graph}
\label{sec:meta}
\noindent \textbf{Metadata.} Documents on the Web are usually accompanied by rich metadata information \cite{zhang2020minimally,mekala2020meta,zhang2021match}. 
To provide a holistic view of documents with metadata, we can construct a heterogeneous information network (HIN) \cite{sun2012mining} to connect documents together.

\begin{definition}{(Heterogeneous Information Network \cite{sun2012mining})} 
An HIN is a graph $G = (\mathcal{V}, \mathcal{E})$ with a node type mapping $\phi: \mathcal{V}\rightarrow\mathcal{T}_\mathcal{V}$ and an edge type mapping $\psi: \mathcal{E}\rightarrow\mathcal{T}_\mathcal{E}$. Either the number of node types $|\mathcal{T}_\mathcal{V}|$ or the number of edge types $|\mathcal{T}_\mathcal{E}|$ is larger than 1.
\label{def:hin}
\end{definition}

As shown in Figure \ref{fig:intro}(a), in our constructed HIN, each document is a node, and each metadata field is described by either a node (e.g., author, venue) or an edge (e.g., reference).

\vspace{1mm}

\noindent \textbf{Meta-Path.} With the constructed HIN, we are able to analyze various relations between documents. Due to network heterogeneity, two documents can be connected via different paths. For example, when two papers share a common author, they can be connected via ``paper--author--paper''; when one paper cites the other, they can also be connected via ``paper$\rightarrow$paper''. To capture the proximity between two nodes from different semantic perspectives, meta-paths \cite{sun2011pathsim} are extensively used in HIN studies.

\begin{figure}[t]
\centering
\includegraphics[width=0.46\textwidth]{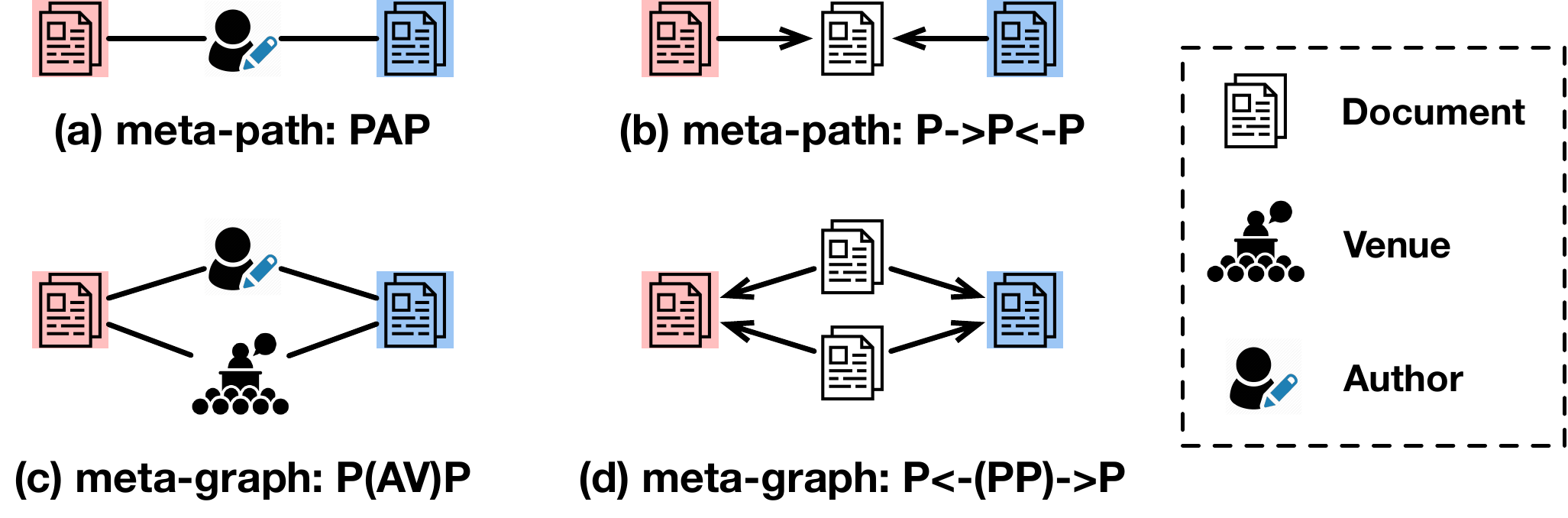}
\vspace{-1em}
\caption{Examples of meta-paths and meta-graphs. Each meta-path/meta-graph describes one type of relation between the red paper and the blue paper.}
\vspace{-1em}
\label{fig:path}
\end{figure}

\begin{definition}{(Meta-Path \cite{sun2011pathsim})} 
A meta-path is a path $\mathcal{M}$ defined on the graph $T_G = (\mathcal{T}_\mathcal{V}, \mathcal{T}_\mathcal{E})$, and is denoted in the form of $\mathcal{M}=V_1\xrightarrow{E_1}V_2\xrightarrow{E_2}\cdots \xrightarrow{E_{m-1}}V_m$, where $V_1,...,V_m$ are node types and $E_1,...,E_{m-1}$ are edge types.
\label{def:metapath}
\end{definition}
Each node is abstracted by its type in a meta-path, and a meta-path describes a composite relation between node types $V_1$ and $V_m$. Figures \ref{fig:path}(a) and \ref{fig:path}(b) show two examples of meta-paths. Following previous studies \cite{sun2011pathsim,dong2017metapath2vec}, we use initial letters to represent node types (e.g., $P$ for paper, $A$ for author) and omit the edge types when there is no ambiguity. The two meta-paths in Figures \ref{fig:path}(a) and \ref{fig:path}(b) can be written as $PAP$ and $P\rightarrow P\leftarrow P$, respectively.\footnote{Following previous studies \cite{sun2011co,sun2011pathsim}, we view $P_1 \rightarrow P_2 \leftarrow P_3$ as a ``directed path'' from $P_1$ to $P_3$ by explaining it as $P_1 \xrightarrow{\rm cites} P_2 \xrightarrow{\rm is\ cited\ by} P_3$. In this way, $P\rightarrow P\leftarrow P$ can be defined as a meta-path according to Definition \ref{def:metapath}. Similarly, we view $PAP$ as a ``directed path'' $P\xrightarrow{\rm writes}A\xrightarrow{\rm is\ written\ by}P$. Using the same explanation, both $P(AV)P$ and $P\leftarrow(PP)\rightarrow P$ in Figure \ref{fig:path} can be viewed as a DAG, thus they are meta-graphs according to Definition \ref{def:metagraph}.}

\vspace{1mm}

\noindent \textbf{Meta-Graph.} In some cases, paths may not be sufficient to capture latent semantics between two nodes. For example, one meta-path cannot describe the relation between two papers that share \textit{at least two} authors. Note that this relation is worth studying because two co-authors can be more informative than a single author when we infer semantic similarities between papers. Allegedly, one researcher may work on multiple topics during his/her career, while the collaboration between two researchers often focuses on a specific direction. Meta-graphs \cite{zhang2018metagraph2vec} are proposed to depict such complex relations in an HIN.

\begin{definition}{(Meta-Graph \cite{zhang2018metagraph2vec})} 
A meta-graph is a directed acyclic graph (DAG) $\mathcal{M}$ defined on $T_G = (\mathcal{T}_\mathcal{V}, \mathcal{T}_\mathcal{E})$. It has a single source node $V_1$ and a single target node $V_m$. Each node in $\mathcal{M}$ is a node type and each edge in $\mathcal{M}$ is an edge type.
\label{def:metagraph}
\end{definition}
Figures \ref{fig:path}(c) and \ref{fig:path}(d) give two examples of meta-graphs. Figure \ref{fig:path}(c) describes two papers sharing the same venue and one common author, while Figure \ref{fig:path}(d) shows two papers both cited by another two shared papers. Similar to the notation of meta-paths, we denote them as $P(AV)P$ and $P\leftarrow(PP)\rightarrow P$.

\vspace{1mm}

\noindent \textbf{Reachability.} We assume that two documents connected via a certain meta-path/meta-graph share similar topics. Formally, we introduce the concept of ``reachable''.
\begin{definition}{(Reachable)}
Given a meta-path/meta-graph $\mathcal{M}$ and two documents $d_1, d_2$, we say $d_2$ is reachable from $d_1$ via $\mathcal{M}$ if and only if we can find a path/DAG $\mathcal{M}_0$ in the HIN such that: (1) $d_1$ is the source node of $\mathcal{M}_0$; (2) $d_2$ is the target node of $\mathcal{M}_0$; (3) when each node in $\mathcal{M}_0$ is abstracted by its node type, $\mathcal{M}_0$ becomes $\mathcal{M}$.
\label{def:reachable}
\end{definition}
We use $d_1 \rightarrow^{\mathcal{M}} d_2$ to denote that $d_2$ is reachable from $d_1$ via $\mathcal{M}$, and we use $\mathcal{N}_{\mathcal{M}}(d_1)$ to denote the set of nodes that are reachable from $d_1$ (i.e., $\mathcal{N}_{\mathcal{M}}(d_1)=\{d_2\ |\ d_1 \rightarrow^{\mathcal{M}} d_2, \ d_2\neq d_1\}$).

\subsection{Problem Definition}
Zero-shot multi-label text classification \cite{wang2019survey} aims to tag each document with labels that are \textit{unseen} during training time but available for prediction. Most previous studies \cite{rios2018few,chalkidis2019large,gupta2021generalized,nam2015predicting} assume that there is a set of \textit{seen} classes, each of which has some annotated documents. Trained on these documents, their proposed text classifiers are expected to transfer the knowledge from seen classes to the prediction of unseen ones.

In this paper, we study a more challenging setting (proposed in \cite{yin2019benchmarking} previously), where all labels are \textit{unseen}. In other words, given the label space $\mathcal{L}$, we do not have any training sample for $l \in \mathcal{L}$. Instead, we assume a large-scale unlabeled corpus $\mathcal{D}$ is given, and each document $d \in \mathcal{D}$ is associated with metadata information. As mentioned in Section \ref{sec:meta}, with such metadata, we can construct an HIN $G = (\mathcal{V}, \mathcal{E})$ to describe the relations between documents. We aim to train a multi-label text classifier $f$ based on both the text information $\mathcal{D}$ and the network information $G$. As an \textit{inductive} task, the classifier $f$ needs to predict its relevant labels given a new document $d \notin \mathcal{D}$.

Since no training data is available to characterize a given label, same as previous studies \cite{rios2018few,yin2019benchmarking}, we assume each label $l$ has some text information to describe its semantics, such as label names $n_l$ \cite{yin2019benchmarking,shen2021taxoclass} and descriptions $s_l$ \cite{chang2020taming,chai2020description}.\footnote{Label names are required in our \textsc{\model} framework, but label descriptions are optional. That being said, \textsc{\model} is still applicable with label names only (i.e., $s_l=\emptyset$).} Examples of such label information have been shown in Figure \ref{fig:label}. To summarize, our task can be formally defined as follows:
\begin{definition}{(Problem Definition)}
Given an unlabeled corpus $\mathcal{D}$ with metadata information $G = (\mathcal{V}, \mathcal{E})$, and a label space $\mathcal{L}$ with label names and descriptions $\{n_l, s_l|l\in \mathcal{L}\}$, our task is to learn a multi-label text classifier $f$ that can map a new document $d \notin \mathcal{D}$ to its relevant labels $\mathcal{L}_d \subseteq \mathcal{L}$.
\label{def:problem}
\end{definition}

\section{The \textsc{\model} Framework}
\subsection{A Two-Stage Framework}
As shown in Figure \ref{fig:intro}(b), in the proposed \textsc{\model} framework, the LMTC problem is formulated as a \underline{\textit{ranking}} task. Specifically, given a new document (i.e., the ``query''), our task is to predict top-ranked labels (i.e., the ``items'') that are relevant to the document. Note that in LMTC, the label space $\mathcal{L}$ (i.e., the ``item pool'') is large. For example, in both the Microsoft Academic and PubMed datasets \cite{zhang2021match}, there are more than 15,000 labels. Given a large item pool, recent ranking approaches are usually pipelined \cite{nogueira2019multi,gao2021rethink}, consisting of a first-stage discrete \textit{retriever} (e.g., BM25 \cite{robertson1994some}) that efficiently generates a set of candidate items followed by a continuous \textit{re-ranker} (e.g., BERT \cite{devlin2019bert}) that selects the most promising items from the candidates. Such design is a natural choice due to the effectiveness-efficiency trade-off among different ranking models: discrete rankers based on lexical matching are faster but less accurate; continuous rankers can perform latent semantic matching but are much slower. 

Following such prevalent approaches, \textsc{\model} adopts a two-stage ranking framework, with a discrete retrieval stage and a continuous re-ranking stage. The major novelty of \textsc{\model} is that, with document metadata information, a new contrastive learning method is developed to significantly improve the \textit{re-ranking} stage performance upon BERT-based models.

\subsection{The Retrieval Stage}
\label{sec:retrieval}
Since the main goal of this paper is to develop a novel contrastive learning framework for re-ranking, we do not aim at a complicated design of the retrieval stage. Therefore, we adopt two simple strategies: \textit{exact name matching} and \textit{sparse retrieval}.

\vspace{1mm}

\noindent \textbf{Exact Name Matching.} Given a document $d$ and a label $l$, if the label name $n_l$ appears in the document text, we add $l$ as a candidate of $d$'s relevant labels.\footnote{On PubMed, as shown in Figure \ref{fig:label}, each label can have more than one name because both ``MeSH heading'' and ``entry term(s)'' are viewed as label names. In this case, we add $l$ as a candidate if \textit{any} of its names appears in the document text.} We use $\mathcal{C}_{\rm exact}(d)$ to denote the set of candidate labels obtained by exact name matching. 

\vspace{1mm}

\noindent \textbf{Sparse Retrieval.} We cannot expect all relevant labels of a document explicitly appear in its text. To increase the recall of our retrieval stage, we adopt BM25 \cite{robertson1994some} to allow \textit{partial} lexical matching between documents and labels. Specifically, we concatenate the name and the description together as the text information $t_l$ of each label (i.e., $t_l = n_l\ ||\ s_l$).\footnote{On PubMed, instead of concatenating all label names into $t_l$, we use the ``MeSH heading'' only as $n_l$, which achieves better performance in experiments.} Then, the score between $d$ and $l$ is calculated as
\begin{equation}
    {\rm BM25}(d, l) = \sum_{w \in d \cap t_l} {\rm IDF}(w) \frac{{\rm TF}(w, t_l)\cdot (k_1+1)}{{\rm TF}(w, t_l)\cdot k_1(1-b+b\frac{|\mathcal{L}|}{avgdl})}.
\end{equation}
Here, $k_1=1.5$ and $b=0.75$ are parameters of BM25; $avgdl = \frac{1}{|\mathcal{L}|}\sum_{l \in \mathcal{L}}|t_l|$ is the average length of label text information. Note that in classification tasks, documents are ``queries'' and labels are ``items'' being ranked.
When the BM25 score between $d$ and $l$ exceeds a certain threshold $\eta$, we add $l$ as a candidate of $d$'s relevant labels. Formally,
\begin{equation}
    \mathcal{C}_{\rm BM25}(d) = \{l\ |\ l \in \mathcal{L},\ {\rm BM25}(d, l) > \eta\}.
\label{eqn:bm25}
\end{equation}

Given a document $d$, its candidate label set $\mathcal{C}(d) = \mathcal{C}_{\rm exact}(d) \cup \mathcal{C}_{\rm BM25}(d)$ (i.e., the union of candidates obtained by exact name matching and by sparse retrieval).

\subsection{The Re-ranking Stage}
\label{sec:rerank}
Encouraged by the success of BERT \cite{devlin2019bert} in a wide range of text mining tasks, we build our re-ranker upon BERT-based pre-trained language models. In general, our proposed re-ranking stage can be instantiated by any variant of BERT (e.g., SciBERT \cite{beltagy2019scibert}, BioBERT \cite{liu2019roberta}, and RoBERTa \cite{lee2020biobert}). In our experiments, since documents from both Microsoft Academic and PubMed are scientific papers, we adopt SciBERT \cite{beltagy2019scibert} as our building block. 

\subsubsection{Bi-Encoder and Cross-Encoder}
To improve the performance of BERT, two architectures are typically used for fine-tuning: Bi-Encoders and Cross-Encoders. Bi-Encoders \cite{reimers2019sentence,cohan2020specter} perform self attention over two text units (e.g., query and item) separately and compute the similarity between their representation vectors at the end. Cross-Encoders \cite{yin2019benchmarking,gao2021rethink}, in contrast, perform self attention \textit{within} as well as \textit{across} two text units at the same time. Below we introduce how to apply these two architectures to our task.

\vspace{1mm}

\noindent \textbf{Bi-Encoder.} Given a document $d$ and a candidate label $l \in \mathcal{C}(d)$ (obtained in the retrieval stage), we use BERT to encode them separately to generate two representation vectors.
\begin{equation}
\bme_d = {\rm BERT}(d), \ \ \ \ \bme_l = {\rm BERT}(t_l).
\label{eqn:bifwd}
\end{equation}
To be specific, given the document text $d$ (resp., the label name and description $t_l$), we use the sequence ``[CLS] $d$ [SEP]'' (resp., ``[CLS] $t_l$ [SEP]'') as the input into BERT and take the output vector of the ``[CLS]'' token from the last layer as the document representation $\bme_d$ (resp., label representation $\bme_l$). The score between $d$ and $l$ is defined as the cosine similarity of their representation vectors.
\begin{equation}
    score(d, l) = \cos(\bme_d, \bme_l).
\label{eqn:biscore}
\end{equation}

\vspace{1mm}

\noindent \textbf{Cross-Encoder.} To better utilize the fully connected attention mechanism of BERT-based models, we can concatenate document and label text information together and encode it using one BERT.
\begin{equation}
\bme_{d || t_l} = {\rm BERT}(d \ || \ t_l).
\label{eqn:crossfwd}
\end{equation}
Here, $(d \ || \ t_l)$ denotes the input sequence ``[CLS] $d$ [SEP] $t_l$ [SEP]''. Again, we take the output vector of the ``[CLS]'' token as $\bme_{d || t_l}$. The score between $d$ and $l$ is then obtained by adding a linear layer upon BERT:
\begin{equation}
    score(d, l) = \bmw^\top\bme_{d || t_l},
\label{eqn:crossscore}
\end{equation}
where $\bmw$ is a trainable vector.

\begin{figure}
\centering
\vspace{-0.5em}
\subfigure[Bi-Encoder]{
\includegraphics[scale=0.35]{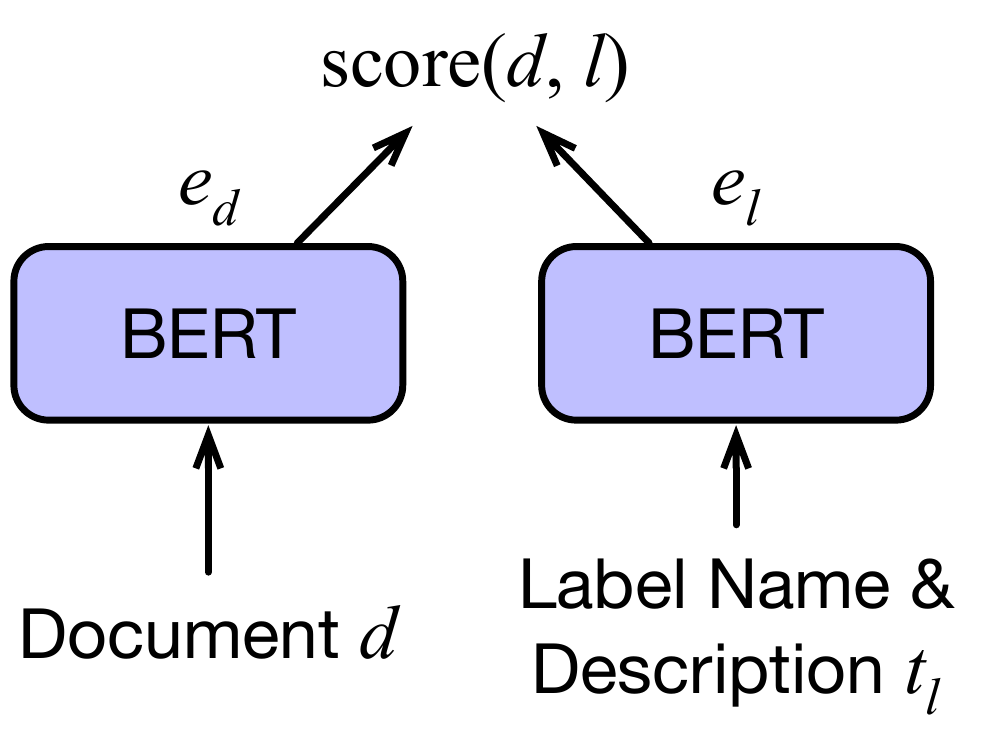}}
\hspace{4mm}
\subfigure[Cross-Encoder]{
\includegraphics[scale=0.35]{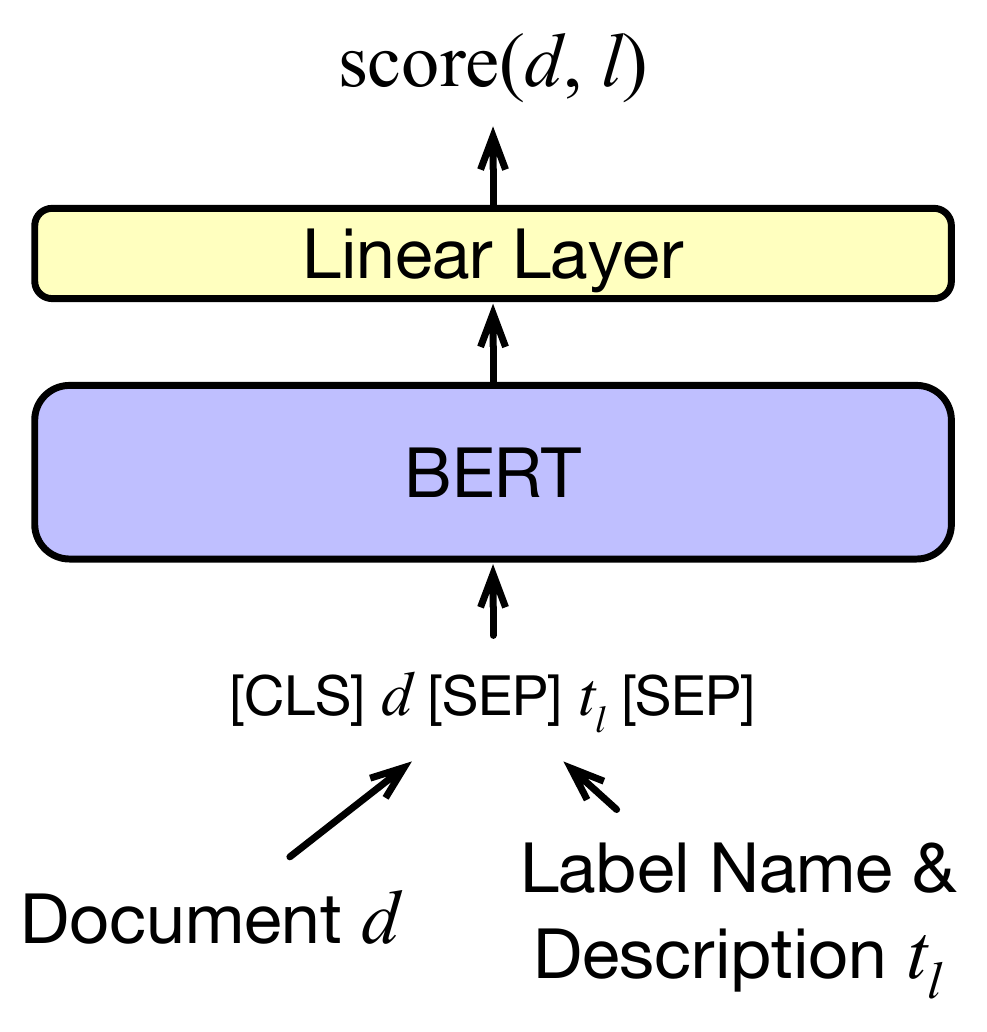}}
\vspace{-1em}
\caption{Two architectures that compute the similarity between a document $d$ and a candidate label $l$.} 
\vspace{-1em}
\label{fig:inference}
\end{figure}

The architectures of Bi-Encoder and Cross-Encoder are illustrated in Figure \ref{fig:inference}.

\subsubsection{Metadata-Induced Contrastive Learning (\textsc{\model})}
Now we aim to fine-tune Bi-Encoder and Cross-Encoder to improve their re-ranking performance. (For Cross-Encoder, especially, we cannot even run it without fine-tuning because $\bmw$ needs to be learned.) If our task were fully supervised, we would have positive document--label training pairs $(d,l)$ indicating $d$ is labeled with $l$, and the training objective would be maximizing $score(d, l)$ for these positive pairs. However, we do not have any annotated documents with the zero-shot setting. In this case, to fine-tune above two architectures, we adopt a \textit{contrastive learning} framework.

Instead of learning ``what is what'', contrastive learning \cite{chen2020simple} tries to learn ``what is similar to what''. In our problem setting, we assume there is a collection of document pairs $(d, d^+)$, where $d$ and $d^+$ are similar to each other, e.g., $d^+$ is reachable from $d$ via a specified meta-path or meta-graph. For each $d$, we can also randomly sample a set of documents $\{d_i^-\}_{i=1}^N$ from the whole corpus $\mathcal{D}$. Contrastive learning aims to learn effective representations by pulling $d$ and $d^+$ together while pushing $d$ and $d_i^-$ apart. Taking Bi-Encoder as an example, we first use BERT to encode all documents.
\begin{equation}
    \bme_d = {\rm BERT}(d), \ \ \ \ \bme_{d^+} = {\rm BERT}(d^+), \ \ \ \ \bme_{d_i^-} = {\rm BERT}(d_i^-).
\label{eqn:bien}
\end{equation}
Following Chen et al.'s seminal work \cite{chen2020simple}, the contrastive loss can be defined as
\begin{equation}
    -\log\frac{\exp(\cos(\bme_d, \bme_{d^+})/\tau)}{\exp(\cos(\bme_d, \bme_{d^+})/\tau)+\sum_{i=1}^N \exp(\cos(\bme_d, \bme_{d_i^-})/\tau)},
\end{equation}
where $\tau$ is a temperature hyperparameter.

Now the problem becomes how to define similar document--document pairs $(d, d^+)$. In Chen et al.'s original paper \cite{chen2020simple}, they focus on learning visual representations, so they take two random transformations (e.g., cropping, distortion, rotation) of the same image as positive pairs. A similar approach has been adopted in learning language representations \cite{wu2020clear,luo2021unsupervised,giorgi2021declutr,yan2021consert}, but transformation techniques become word insertion, deletion, substitution, reordering \cite{wei2019eda}, and back translation \cite{xie2020unsupervised}. 

Instead of using those purely text-based techniques, we propose a simple but novel approach based on \textit{document metadata}. That is, given a meta-path or a meta-graph $\mathcal{M}$, we define $(d, d^+)$ as a similar document--document pair if and only if $d^+$ is reachable 
from $d$ via $\mathcal{M}$ (i.e., $d^+ \in \mathcal{N}_{\mathcal{M}}(d)$, Definition \ref{def:reachable}).

\begin{figure}
\centering
\vspace{-0.5em}
\subfigure[Bi-Encoder fine-tuning]{
\includegraphics[scale=0.35]{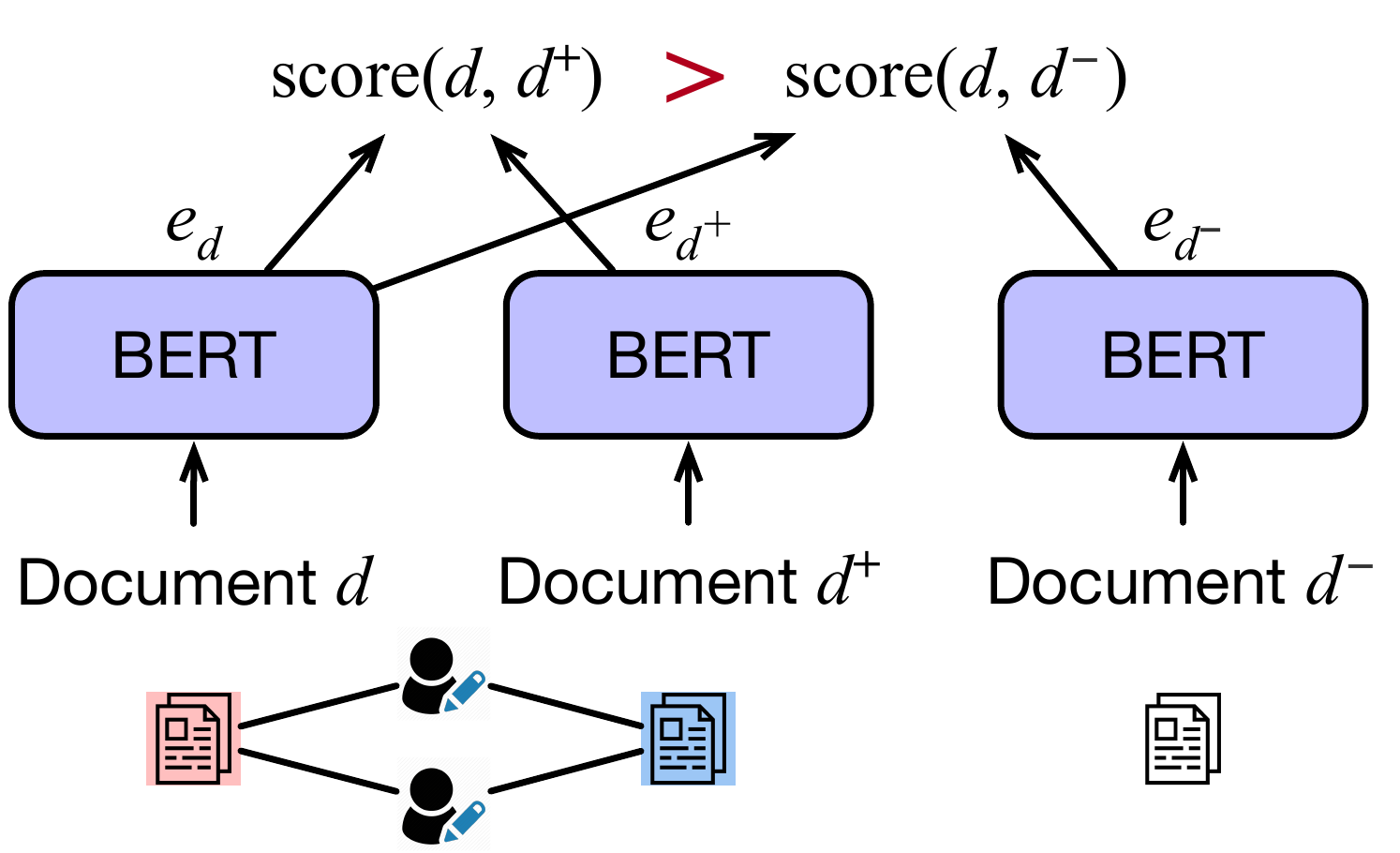}} \\
\vspace{-0.5em}
\subfigure[Cross-Encoder fine-tuning]{
\includegraphics[scale=0.35]{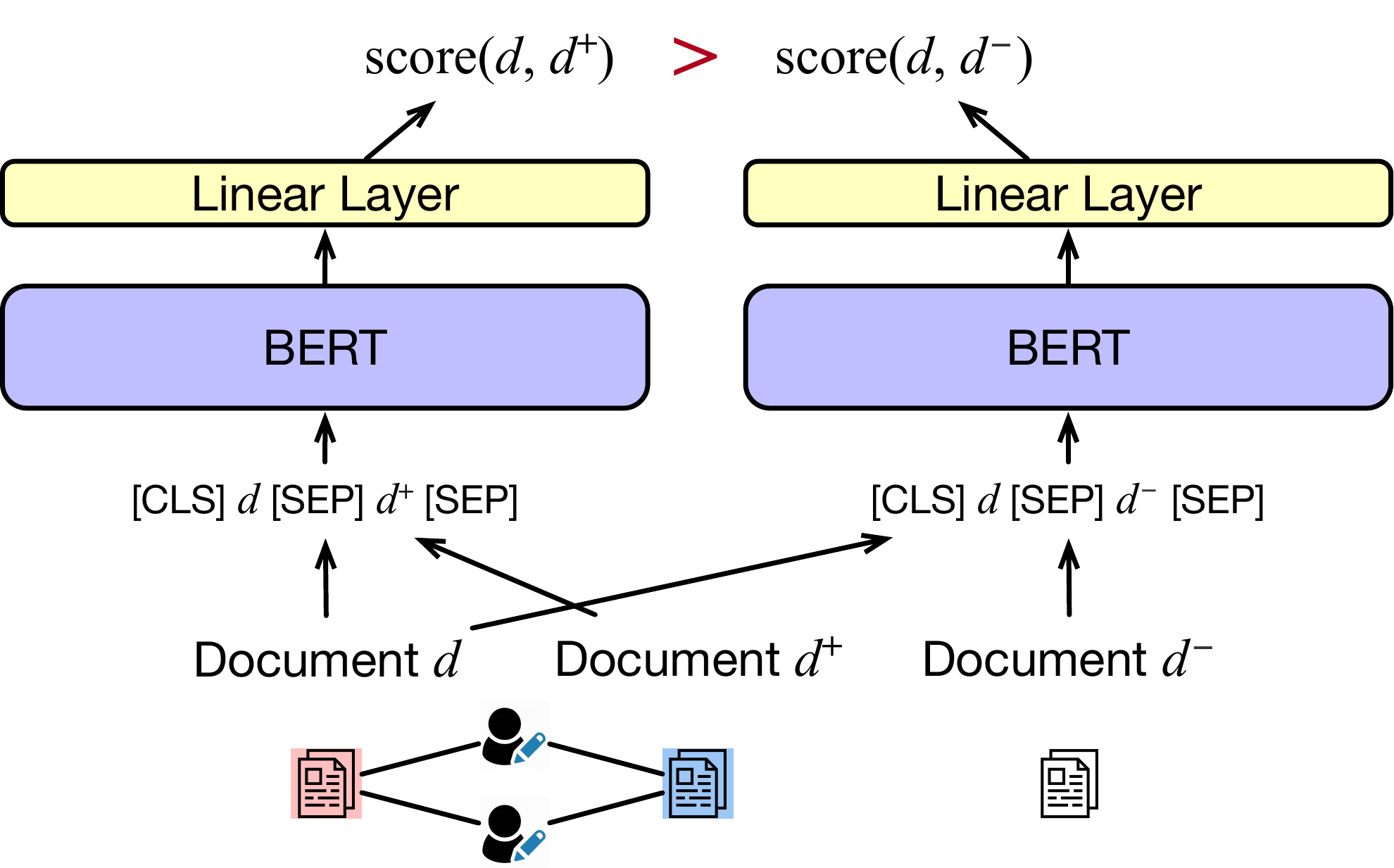}}
\vspace{-1em}
\caption{Metadata-induced contrastive learning to fine-tune Bi-Encoder and Cross-Encoder using $\mathcal{M}=P(AA)P$.} 
\vspace{-1em}
\label{fig:train}
\end{figure}

Formally, for \textit{Bi-Encoder}, the metadata-induced contrastive loss is defined as

\begin{equation}
\small
    \mathcal{J}_{\rm Bi} =
    \mathop{\mathbb{E}}_{\substack{d^+ \in \mathcal{N}_{\mathcal{M}}(d) \\ d_i^-\sim \mathcal{D}}}  \bigg[-\log\frac{\exp(\cos(\bme_d, \bme_{d^+})/\tau)}{\exp(\cos(\bme_d, \bme_{d^+})/\tau)+\sum_{i=1}^N \exp(\cos(\bme_d, \bme_{d_i^-})/\tau)}\bigg].
\label{eqn:bi}
\end{equation}

Similarly, for \textit{Cross-Encoder}, we first compute $score(d, d^+)$ and $score(d, d_i^-)$.
\begin{equation}
\begin{split}
    \bme_{d || d^+} = {\rm BERT}(d \ || \ d^+), \ \ \ &\ \ \ \ \bme_{d || d_i^-} = {\rm BERT}(d \ || \ d_i^-), \\
    score(d, d^+) = \bmw^\top\bme_{d || d^+}, \ \ \ &\ \ \ \ score(d, d_i^-) = \bmw^\top \bme_{d || d_i^-}.
\end{split}
\label{eqn:crossen}
\end{equation}
Then, the metadata-induced contrastive loss is
\begin{equation}
\small
    \mathcal{J}_{\rm Cross} =
    \mathop{\mathbb{E}}_{\substack{d^+ \in \mathcal{N}_{\mathcal{M}}(d) \\ d_i^-\sim \mathcal{D}}}  \bigg[-\log\frac{\exp(score(d, d^+))}{\exp(score(d, d^+))+\sum_{i=1}^N \exp(score(d, d_i^-))}\bigg].
\label{eqn:cross}
\end{equation}
The BERT model is thus fine-tuned by minimizing the contrastive loss in Eq. (\ref{eqn:bi}) or (\ref{eqn:cross}). Figure \ref{fig:train} illustrates the fine-tuning process of both Bi-Encoder and Cross-Encoder using $\mathcal{M}=P(AA)P$.

\vspace{1mm}

Due to the space constraint, the training and inference procedures of the \textsc{\model} framework are formally summarized in Appendix \ref{sec:appen_sum}; the optimization details of \textsc{\model} (i.e., Eqs. (\ref{eqn:bi}) and (\ref{eqn:cross})) are provided in Appendix \ref{sec:appen_impl}.

\section{Experiments}
\subsection{Setup}
\noindent \textbf{Datasets.} Given the task of metadata-aware LMTC, following \cite{zhang2021match}, we perform evaluation on two large-scale datasets.
\begin{itemize}[leftmargin=*]
\item \textbf{MAG-CS \cite{wang2020microsoft}.} The Microsoft Academic Graph (MAG) has a web-scale collection of scientific papers from various fields. In \cite{zhang2021match}, 705,407 MAG papers published at 105 top CS conferences from 1990 to 2020 are selected to form a dataset with 15,808 labels.\footnote{Originally, there were 15,809 labels in MAG-CS, but the label ``\texttt{Computer Science}'' is removed from all papers because it is trivial to predict.}
\item \textbf{PubMed \cite{lu2011pubmed}.} PubMed has a web-scale collection of biomedical literature from MEDLINE, life science journals, and online books. In \cite{zhang2021match}, 898,546 PubMed papers published in 150 top medicine journals from 2010 to 2020 are selected to form a dataset with 17,963 labels (i.e., MeSH terms \cite{coletti2001medical}). 
\end{itemize}

Under the fully supervised setting, Zhang et al. \cite{zhang2021match} split both datasets into training, validation, and testing sets. In this paper, we focus on the zero-shot setting. Therefore, we combine their training and validation sets together as our \underline{\textit{unlabeled}} input corpus $\mathcal{D}$ (that being said, we do \underline{\textit{not}} know the labels of these documents, and we only utilize their text and metadata information). We use their testing set as our testing documents $d \notin \mathcal{D}$. Dataset statistics are briefly listed in Table \ref{tab:data}. More details are in Appendix \ref{sec:appen_data}.

\begin{table}[!h]
\small
\centering
\vspace{-0.5em}
\caption{Dataset statistics.}
\vspace{-1em}
\scalebox{0.88}{
\begin{tabular}{c|ccccc}
\hline
Dataset & \begin{tabular}[c]{@{}c@{}}\#Training\\ (Unlabeled)\end{tabular} & \#Testing & \#Labels & Labels/Doc & Words/Doc \\ \hline
MAG-CS \cite{wang2020microsoft} & 634,874                                                          & 70,533    & 15,808   & 5.59   & 126.55    \\
PubMed \cite{lu2011pubmed}  & 808,692                                                          & 89,854    & 17,963   & 7.80  & 199.14     \\ \hline
\end{tabular}
}
\vspace{-0.5em}
\label{tab:data}
\end{table}

\begin{table*}[]
\small
\caption{P@$k$ and NDCG@$k$ scores of compared algorithms on MAG-CS and PubMed. Bold: the highest score of zero-shot approaches. *: \textsc{\model} (Cross-Encoder, $P\rightarrow P\leftarrow P$) is significantly better than this algorithm with p-value $< 0.05$. **: \textsc{\model} (Cross-Encoder, $P\rightarrow P\leftarrow P$) is significantly better than this algorithm with p-value $< 0.01$.}
\vspace{-0.5em}
\scalebox{0.86}{
\begin{tabular}{l|c|ccccc|ccccc}
\hline
& \multirow{2}{*}{\textbf{Algorithm}} & \multicolumn{5}{c|}{\textbf{MAG-CS \cite{wang2020microsoft}}}                                                             & \multicolumn{5}{c}{\textbf{PubMed \cite{lu2011pubmed}}}                                                              \\ \cline{3-12} 
&                            & P@1             & P@3             & P@5             & NDCG@3          & NDCG@5          & P@1             & P@3             & P@5             & NDCG@3          & NDCG@5          \\ \hline
{\multirow{10}{*}{\rotatebox[origin=c]{90}{Zero-shot}}} &
Doc2Vec \cite{mikolov2013distributed}                   & 0.5697**          & 0.4613**          & 0.3814**          & 0.5043**          & 0.4719**          & 0.3888**          & 0.3283**          & 0.2859**          & 0.3463**          & 0.3252**          \\
& SciBERT \cite{beltagy2019scibert}                   & 0.6440**          & 0.5030**          & 0.4011**          & 0.5545**          & 0.5061**          & 0.4427**          & 0.3572**          & 0.3031**          & 0.3809**          & 0.3510**          \\
& ZeroShot-Entail \cite{yin2019benchmarking}           & 0.6649**          & 0.5003**          & 0.3959**          & 0.5570**          & 0.5057**          & 0.5275**          & 0.4021          & \textbf{0.3299} & 0.4352          & \textbf{0.3913} \\
& SPECTER \cite{cohan2020specter}                    & 0.7107**          & 0.5381**          & 0.4184**          & 0.5979**          & 0.5365**          & 0.5286**          & 0.3923**          & 0.3181**          & 0.4273**          & 0.3815**          \\
& EDA \cite{wei2019eda}                       & 0.6442**          & 0.4939**          & 0.3948**          & 0.5471**          & 0.5000**          & 0.4919          & 0.3754*          & 0.3101*          & 0.4058*          & 0.3667*          \\
& UDA \cite{xie2020unsupervised}                       & 0.6291**          & 0.4848**          & 0.3897**          & 0.5362**          & 0.4918**          & 0.4795**          & 0.3696**          & 0.3067**          & 0.3986**          & 0.3614**          \\ \cline{2-12}
& \textsc{\model} (Bi-Encoder, $P\rightarrow P\leftarrow P$)            & 0.7062*          & 0.5369*          & 0.4184*          & 0.5960*          & 0.5355*          & 0.5124**          & 0.3869*          & 0.3172*          & 0.4196*          & 0.3774*          \\
& \textsc{\model} (Bi-Encoder, $P\leftarrow (PP)\rightarrow P$)            & 0.7050*          & 0.5344*          & 0.4161*          & 0.5937*          & 0.5331*          & 0.5198**          & 0.3876*          & 0.3172*          & 0.4215*          & 0.3786*          \\
& \textsc{\model} (Cross-Encoder, $P\rightarrow P\leftarrow P$)         & \textbf{0.7177} & \textbf{0.5444} & \textbf{0.4219} & \textbf{0.6048} & \textbf{0.5415} & \textbf{0.5412} & \textbf{0.4036} & 0.3257          & \textbf{0.4391} & 0.3906          \\
& \textsc{\model} (Cross-Encoder, $P\leftarrow (PP)\rightarrow P$)         & 0.7061          & 0.5376          & 0.4187          & 0.5964          & 0.5357          & 0.5218          & 0.3911          & 0.3172*          & 0.4249          & 0.3794          \\ \hline \hline
{\multirow{4}{*}{\rotatebox[origin=c]{90}{Supervised}}} &
MATCH \cite{zhang2021match} (10K Training)            & 0.4423**          & 0.2851**          & 0.2152**          & 0.3375**          & 0.3003**          & 0.6915          & 0.3869*          & 0.2785**          & 0.4649          & 0.3896          \\
& MATCH \cite{zhang2021match} (50K Training)            & 0.6215**          & 0.4280**          & 0.3269**          & 0.4987**          & 0.4489**          & 0.7701          & 0.4716          & 0.3585          & 0.5497          & 0.4750          \\
& MATCH \cite{zhang2021match} (100K Training)           & 0.8321          & 0.6520          & 0.5142          & 0.7342          & 0.6761          &   0.8286              &     0.5680            &    0.4410             &      0.6405           &    0.5626             \\ 
& MATCH \cite{zhang2021match} (Full, 560K+ Training)           & 0.9114          & 0.7634          & 0.6312          & 0.8486          & 0.8076          &   0.9151              &     0.7425            &    0.6104             &      0.8001           &    0.7310             \\ \hline
\end{tabular}
}
\label{tab:patk}
\end{table*}

\noindent \textbf{Compared Methods.} We evaluate \textsc{\model} against a variety of baseline methods using text embedding, pre-trained language models, and text-based contrastive learning. Since the major technical contribution of \textsc{\model} is in the re-ranking stage, all the baselines below are used as re-rankers after the same retrieval stage proposed in Section \ref{sec:retrieval}.
\begin{itemize}[leftmargin=*]
\item \textbf{Doc2Vec \cite{le2014distributed}} is a text embedding method. We use it to embed documents and labels into a shared semantic space according to their text information. Then, for each document, we rank all candidate labels according to their cosine similarity with the document in the embedding space.
\item \textbf{SciBERT \cite{beltagy2019scibert}} is a BERT-based language model pre-trained on a large set of computer science and biomedical papers. Taking it as a baseline, we do not perform fine-tuning, so it can only be used in Bi-Encoder. (In Cross-Encoder, the linear layer is not pre-trained, thus cannot be used without fine-tuning.)
\item \textbf{ZeroShot-Entail \cite{yin2019benchmarking}} is a pre-trained language model for zero-shot text classification. It is a textual entailment model that predicts to what extent a document (as the premise) can entail the template ``\textit{this document is about \{label\_name\}}'' (as the hypothesis). To make this method more competitive, we change its internal \texttt{BERT-base-uncased} model to \texttt{RoBERTa-large-mnli}.
\item \textbf{SPECTER \cite{cohan2020specter}} is a pre-trained language model for scientific documents that leverages paper citation information. It is built upon SciBERT and takes citation prediction as the pre-training objective. We use it in the Bi-Encoder architecture without fine-tuning.
\item \textbf{EDA \cite{wei2019eda}} is a text data augmentation method. Given a document, it proposes four simple operations -- synonym replacement, random insertion, random swap, and random deletion -- to create a new artificial document. We view the original document and the new one as a positive document--document pair and use all these pairs to perform contrastive learning to fine-tune SciBERT.
\item \textbf{UDA \cite{xie2020unsupervised}} is another text data augmentation method. It performs back translation and TF-IDF word replacement to generate new documents that are similar to the original one. We use these pairs to perform contrastive learning to fine-tune SciBERT. Both EDA and UDA can be leveraged to fine-tune a Bi-Encoder or a Cross-Encoder, and we report the higher performance between the two architectures.
\item \textbf{\textsc{\model}} is our proposed framework. We study the performance of 10 meta-paths/meta-graphs $\{P\rightarrow P, P\leftarrow P, PAP, PVP, P\rightarrow P\leftarrow P, P\leftarrow P\rightarrow P, P(AA)P, P(AV)P, P\rightarrow (PP)\leftarrow P, P\leftarrow (PP)\rightarrow P\}$ when fine-tuning Bi-Encoder and Cross-Encoder. We choose \textbf{SciBERT} as our base model to be fine-tuned.
\end{itemize}

\noindent We also report the performance of a fully supervised method for reference.
\begin{itemize}[leftmargin=*]
\item \textbf{MATCH \cite{zhang2021match}} is the state-of-the-art supervised approach for metadata-aware multi-label text classification. Because we do not consider label hierarchy in our problem setting, we report the performance of MATCH-NoHierarchy with various sizes of training data for comparison.
\end{itemize}

\vspace{1mm}

\noindent \textbf{Evaluation Metrics.} Following the commonly used evaluation  on multi-label text classification \cite{liu2017deep,you2019attentionxml,zhang2021match}, we adopt two rank-based metrics: P@$k$ and NDCG@$k$, where $k=1,3,5$. For a document $d$, let $\bmy_d \in \{0,1\}^{|\mathcal{L}|}$ be its ground truth label vector and ${\rm rank}(i)$ be the index of the $i$-th highest predicted label according to the re-ranker.
\begin{equation}
\small
\begin{split}
    {\rm P@}k &= \frac{1}{k} \sum_{i = 1}^k y_{d, {\rm rank}(i)}. \\
    {\rm DCG@}k = \sum_{i=1}^k \frac{y_{d, {\rm rank}(i)}}{\log(i+1)}, \ \ \ &{\rm NDCG@}k = \frac{{\rm DCG@}k}{\sum_{i=1}^{\min(k, ||\bmy_d||_0)}\frac{1}{\log(i+1)}}. \notag
\end{split}
\end{equation}

\begin{table*}[]
\small
\caption{PSP@$k$ and PSN@$k$ scores of compared algorithms on MAG-CS and PubMed. Bold, *, and **: the same meaning as in Table \ref{tab:patk}. We also show the ratio $\rm PSP@1/P@1$. The higher ${\rm PSP@}k/{\rm P@}k$ is, the more infrequent the correctly predicted labels are.}
\vspace{-0.5em}
\scalebox{0.86}{
\begin{tabular}{l|c|cccccc|cccccc}
\hline
& \multirow{2}{*}{\textbf{Algorithm}} & \multicolumn{6}{c|}{\textbf{MAG-CS \cite{wang2020microsoft}}}                                                             & \multicolumn{6}{c}{\textbf{PubMed \cite{lu2011pubmed}}}                                                              \\ \cline{3-14} 
&                           & PSP@1             & PSP@3             & PSP@5             & PSN@3          & PSN@5    &  $\frac{\rm PSP@1}{\rm P@1}$    & PSP@1             & PSP@3             & PSP@5             & PSN@3          & PSN@5     &  $\frac{\rm PSP@1}{\rm P@1}$     \\ \hline
{\multirow{10}{*}{\rotatebox[origin=c]{90}{Zero-shot}}} &
Doc2Vec \cite{mikolov2013distributed}                   & 0.4287**          & 0.4623**          & 0.4656**          & 0.4450**          & 0.4425**     & 0.75     & 0.2717**          & 0.2948**          & 0.3029**          & 0.2856**          & 0.2879**  & 0.70        \\
& SciBERT \cite{beltagy2019scibert}                   & 0.4668**          & 0.4958**          & 0.4843**          & 0.4788**          & 0.4667**      & 0.72    & 0.3149**          & 0.3231**          & 0.3221**          & 0.3174**          & 0.3131**  & 0.71        \\
& ZeroShot-Entail \cite{yin2019benchmarking}           & 0.4796**          & 0.4892**          & 0.4759**          & 0.4777**          & 0.4644**     & 0.72     & 0.3617**          & 0.3498**          & 0.3389** & 0.3492**          & 0.3378**  & 0.69 \\
& SPECTER \cite{cohan2020specter}                    & 0.5304          & 0.5334*         & 0.5059*          & 0.5223          & 0.4988*   & 0.75       & 0.3907**          & 0.3638**          & 0.3442**          & 0.3666**          & 0.3489**    & 0.74      \\
& EDA \cite{wei2019eda}                       & 0.4916**          & 0.4968**          & 0.4821**          & 0.4859**          & 0.4708**     & 0.76     & 0.3572*          & 0.3451*          & 0.3334*          & 0.3442*          & 0.3322*       & 0.73   \\
& UDA \cite{xie2020unsupervised}                       & 0.4850**          & 0.4907**          & 0.4771**          & 0.4797**          & 0.4654**    & 0.77      & 0.3547**          & 0.3423**          & 0.3311**          & 0.3416**          & 0.3298**    & 0.74      \\ \cline{2-14}
& \textsc{\model} (Bi-Encoder, $P\rightarrow P\leftarrow P$)            & 0.5176          & 0.5311          & 0.5065          & 0.5175          & 0.4963   & 0.73       & 0.3676**          & 0.3559**          & 0.3423*          & 0.3550**          & 0.3418**       & 0.72   \\
& \textsc{\model} (Bi-Encoder, $P\leftarrow (PP)\rightarrow P$)            & 0.5160          & 0.5281          & 0.5037          & 0.5150          & 0.4940   & 0.73       & 0.3780**          & 0.3589*          & 0.3423*          & 0.3597**          & 0.3450**       & 0.73   \\
& \textsc{\model} (Cross-Encoder, $P\rightarrow P\leftarrow P$)         & \textbf{0.5375} & \textbf{0.5415} & \textbf{0.5118} & \textbf{0.5302} & \textbf{0.5052} & 0.75 & \textbf{0.4105} & \textbf{0.3807} & \textbf{0.3558}          & \textbf{0.3841} & \textbf{0.3625}   & 0.76       \\
& \textsc{\model} (Cross-Encoder, $P\leftarrow (PP)\rightarrow P$)         & 0.5326          & 0.5363          & 0.5087          & 0.5249          & 0.5013     & 0.75     & 0.3871          & 0.3664          & 0.3462          & 0.3677          & 0.3496     & 0.74     \\ \hline \hline
{\multirow{5}{*}{\rotatebox[origin=c]{90}{Supervised}}} & 
MATCH \cite{zhang2021match} (10K Training)            & 0.1978**          & 0.1807**          & 0.1712**          & 0.1850**          & 0.1764**   & 0.45       & 0.2840**          & 0.2138**          & 0.1870**          & 0.2332**          & 0.2139**     & 0.41     \\
& MATCH \cite{zhang2021match} (50K Training)            & 0.2854**          & 0.2830**          & 0.2738**          & 0.2838**          & 0.2780**  & 0.46        & 0.3201**          & 0.2715**          & 0.2532**          & 0.2848**          & 0.2713**    & 0.42      \\
& MATCH \cite{zhang2021match} (100K Training)            & 0.4271**          & 0.4750**          & 0.4737*          & 0.4624**          & 0.4635**  & 0.51        & 0.3576**          & 0.3579**          & 0.3456*          & 0.3584**          & 0.3507**     & 0.43     \\
& MATCH \cite{zhang2021match} (200K Training)            & 0.4695**          & 0.5401          & 0.5530          & 0.5217          & 0.5325     & 0.54     & 0.3732**          & 0.3988          & 0.3905          & 0.3913          & 0.3882      & 0.44    \\
& MATCH \cite{zhang2021match} (Full, 560K+ Training)           & 0.5501          & 0.6397          & 0.6627          & 0.6171          & 0.6345    & 0.60      &   0.4371              &    0.5188             &   0.5200              &      0.4978           &     0.5011     & 0.48       \\ \hline
\end{tabular}
}
\label{tab:pspk}
\end{table*}

\subsection{Performance Comparison}
Table \ref{tab:patk} shows P@$k$ and NDCG@$k$ scores of compared algorithms on MAG-CS and PubMed. We run each experiment three times with the average score reported. (SciBERT, Zero-Shot-Entail, and SPECTER are deterministic according to our usage, so we run them only once.) To show statistical significance, we conduct two-tailed unpaired t-tests to compare the best performed \textsc{\model} model and other approaches including MATCH. (When comparing \textsc{\model} with three deterministic approaches, we conduct two-tailed Z-tests instead.) The significance level of each result is marked in Table \ref{tab:patk}. For \textsc{\model}, we show the performance of one meta-path $P\rightarrow P\leftarrow P$ and one meta-graph $P\leftarrow (PP)\rightarrow P$ here. The performance of other meta-paths/meta-graphs will be presented in Table \ref{tab:path} and discussed in Section \ref{sec:path}.

From Table \ref{tab:patk}, we observe that: (1) \textsc{\model} (Cross-Encoder, $P\rightarrow P\leftarrow P$) significantly outperforms all zero-shot baselines in most cases, except that it is slightly worse than ZeroShot-Entail on PubMed in terms of P@5 and NDCG@5. (2) On MAG-CS, \textsc{\model} (Cross-Encoder, $P\rightarrow P\leftarrow P$) performs significantly better than the supervised MATCH model with 50K labeled training data. On PubMed, \textsc{\model} can be competitive with MATCH trained on more than 10K annotated documents in terms of P@3, P@5, and NDCG@5. (3) Using purely text-based augmentation approaches (i.e., EDA and UDA) to perform contrastive learning is not consistently beneficial. In fact, EDA and UDA perform even worse than unfine-tuned SciBERT on MAG-CS. In contrast, our proposed metadata-induced contrastive learning consistently boosts the performance of SciBERT, and the improvements are much more significant than those of text-based contrastive learning. (4) For both $P\rightarrow P\leftarrow P$ and $P\leftarrow (PP)\rightarrow P$, Cross-Encoder performs better than Bi-Encoder within the \textsc{\model} framework. In Section \ref{sec:path}, we will show that this observation is generalizable to most meta-paths and meta-graphs we use.

\subsection{Performance on Tail Labels}
Tail labels refer to those labels relevant to only a few documents in the dataset. They are usually more fine-grained and informative than head labels (i.e., frequent ones). However, predicting tail labels is less ``rewarding'' for models to achieve high P@$k$ and NDCG@$k$ scores. Therefore, new scoring functions are designed to promote infrequent label prediction by giving the model a higher ``reward'' when it predicts a tail label correctly. Propensity-scored P@$k$ (PSP@$k$) and propensity-scored NDCG@$k$ (PSNDCG@$k$, abbreviated to PSN@$k$ in this paper) are thus proposed in \cite{jain2016extreme} and widely used in LMTC evaluation \cite{you2019attentionxml,saini2021galaxc,mittal2021eclare,wei2021towards,gupta2021generalized}. PSP@$k$ and PSN@$k$ are defined as follows.\footnote{When reporting PSP@$k$ and PSN@$k$, previous studies \cite{you2019attentionxml,saini2021galaxc,mittal2021eclare,wei2021towards,gupta2021generalized} normalize the original PSP@$k$ and PSN@$k$ scores by their maximum possible values (just like how DCG@$k$ is normalized to NDCG@$k$). Following these studies, we perform the same normalization in our calculation.}
\begin{equation}
\small
\begin{split}
    \frac{1}{p_l} = 1+C(N_l+B)^{-A}, \ \ \ {\rm PSP}&{@}k = \frac{1}{k} \sum_{i = 1}^k \frac{y_{d, {\rm rank}(i)}}{p_{d, {\rm rank}(i)}}. \\
    {\rm PSDCG@}k = \sum_{i=1}^k \frac{y_{d, {\rm rank}(i)}}{p_{d, {\rm rank}(i)}\log(i+1)}, \ \ \ &{\rm PSN@}k = \frac{{\rm PSDCG@}k}{\sum_{i=1}^{\min(k, ||\bmy_d||_0)}\frac{1}{\log(i+1)}}. \notag
\end{split}
\end{equation}
Here, $\frac{1}{p_l}$ is the ``reward'' of predicting the label $l$ correctly; $N_l$ is the number of documents relevant to $l$ in the training set. Following previously established parameter values \cite{jain2016extreme,you2019attentionxml,wei2021towards}, we set $A=0.55$, $B=1.5$, and $C=(\log |\mathcal{D}|-1)(B+1)^A$. Therefore, the less frequent a label is, the higher reward one can get by predicting it correctly. Table \ref{tab:pspk} shows the PSP@$k$ and PSN@$k$ scores of compared methods.

As shown in Table \ref{tab:pspk}, when we use PSP@$k$ and PSN@$k$ as evaluation metrics, \textsc{\model} becomes more powerful. \textsc{\model} (Cross-Encoder, $P\rightarrow P\leftarrow P$) significantly outperforms all zero-shot baselines, and it is on par with MATCH trained on 100K--200K labeled documents. According to the definition of PSP@$k$ and P@$k$, the ratio ${\rm PSP@}k/{\rm P@}k$ reflects the average ``reward'' a model gets from its correctly predicted labels. The higher ${\rm PSP@}k/{\rm P@}k$ is, the more infrequent the correctly predicted labels are. We show $\rm PSP@1/P@1$ in Table \ref{tab:pspk}. We observe that labels predicted by \textsc{\model} (and all other zero-shot methods) are much more infrequent than labels predicted by the supervised MATCH model. The reason could be that zero-shot methods cannot see any labeled data during training, thus they get no hints of frequent labels and are not biased towards head labels. This helps alleviate the deteriorated performance of supervised models on long-tailed labels as observed in \cite{wei2018does,wei2021towards}.

\subsection{Effect of Meta-Path and Meta-Graph}
\label{sec:path}
\begin{table*}[]
\small
\caption{P@$k$ and NDCG@$k$ scores of \textsc{\model} using different meta-paths/meta-graphs. Bold: the best model. *: significantly worse than the best model with p-value $< 0.05$. **: significantly worse than the best model with p-value $< 0.01$. All meta-paths and meta-graphs, except $PVP$, can improve the classification performance upon unfine-tuned SciBERT.}
\vspace{-0.5em}
\scalebox{0.86}{
\begin{tabular}{c|ccccc|ccccc}
\hline
\multirow{2}{*}{\textbf{Algorithm}} & \multicolumn{5}{c|}{\textbf{MAG-CS \cite{wang2020microsoft}}}                                                    & \multicolumn{5}{c}{\textbf{PubMed \cite{lu2011pubmed}}}                                                     \\ \cline{2-11} 
                                    & P@1             & P@3             & P@5             & NDCG@3          & NDCG@5          & P@1             & P@3             & P@5             & NDCG@3          & NDCG@5          \\ \hline
Unfine-tuned SciBERT                   & 0.6599**          & 0.5117**          & 0.4056**          & 0.5651**          & 0.5136**          & 0.4371**          & 0.3544**          & 0.3014**          & 0.3775**          & 0.3485**          \\ \hline
\textsc{\model} (Bi-Encoder, $PAP$)                                 & 0.6877**        & 0.5285**        & 0.4143**        & 0.5852**        & 0.5280**        & 0.4974**        & 0.3818**        & 0.3154*         & 0.4122**        & 0.3727**        \\
\textsc{\model} (Bi-Encoder, $PVP$)                                 & 0.6589**        & 0.5123**        & 0.4063**        & 0.5656**        & 0.5145**        & 0.4440**        & 0.3507**        & 0.2966**        & 0.3761**        & 0.3458**        \\
\textsc{\model} (Bi-Encoder, $P\rightarrow P$)                                  & 0.7094          & 0.5391          & 0.4190          & 0.5982          & 0.5367          & 0.5200*         & 0.3903*         & 0.3195          & 0.4240*         & 0.3808*         \\
\textsc{\model} (Bi-Encoder, $P\leftarrow P$)                                 & 0.7095*         & 0.5374*         & 0.4178*         & 0.5970*         & 0.5356*         & 0.5195**        & 0.3905*         & 0.3192          & 0.4240*         & 0.3806*         \\
\textsc{\model} (Bi-Encoder, $P\rightarrow P\leftarrow P$)                                & 0.7062*         & 0.5369*         & 0.4184*         & 0.5960*         & 0.5355*         & 0.5124**        & 0.3869*         & 0.3172*         & 0.4196*         & 0.3774*         \\
\textsc{\model} (Bi-Encoder, $P\leftarrow P\rightarrow P$)                                 & 0.7039*         & 0.5379*         & 0.4187*         & 0.5963*         & 0.5356*         & 0.5174**        & 0.3886*         & 0.3187*         & 0.4220*         & 0.3795*         \\
\textsc{\model} (Bi-Encoder, $P(AA)P$)                                & 0.6873**        & 0.5272**        & 0.4130**        & 0.5840**        & 0.5269**        & 0.4963**        & 0.3794**        & 0.3139**        & 0.4101**        & 0.3711**        \\
\textsc{\model} (Bi-Encoder, $P(AV)P$)                                & 0.6832**        & 0.5263**        & 0.4135**        & 0.5823**        & 0.5263**        & 0.4894**        & 0.3743**        & 0.3099**        & 0.4045**        & 0.3664**        \\
\textsc{\model} (Bi-Encoder, $P\rightarrow (PP)\leftarrow P$)                               & 0.7015**        & 0.5334**        & 0.4160**        & 0.5920**        & 0.5322**        & 0.5163**        & 0.3879*         & 0.3172*         & 0.4211*         & 0.3781*         \\
\textsc{\model} (Bi-Encoder, $P\leftarrow (PP)\rightarrow P$)                                 & 0.7050*         & 0.5344*         & 0.4161*         & 0.5937*         & 0.5331*         & 0.5198**        & 0.3876*         & 0.3172*         & 0.4215*         & 0.3786*         \\ \hline
\textsc{\model} (Cross-Encoder, $PAP$)                                & 0.7034*         & 0.5355          & 0.4168          & 0.5943          & 0.5337          & 0.5212**        & 0.3921*         & 0.3207          & 0.4255*         & 0.3818*         \\
\textsc{\model} (Cross-Encoder, $PVP$)                                 & 0.6720*         & 0.5203*         & 0.4103*         & 0.5750*         & 0.5210*         & 0.4668**        & 0.3633**        & 0.3051**        & 0.3908**        & 0.3574**        \\
\textsc{\model} (Cross-Encoder, $P\rightarrow P$)                                  & 0.7033*         & 0.5391          & 0.4201          & 0.5971*         & 0.5365*         & 0.5266          & 0.3946          & 0.3207          & 0.4286          & 0.3830          \\
\textsc{\model} (Cross-Encoder, $P\leftarrow P$)                                  & 0.7169          & 0.5430          & 0.4214          & 0.6033          & 0.5406          & 0.5265          & 0.3924          & 0.3186          & 0.4268          & 0.3811          \\
\textsc{\model} (Cross-Encoder, $P\rightarrow P\leftarrow P$)                                 & \textbf{0.7177} & \textbf{0.5444} & \textbf{0.4219} & \textbf{0.6048} & \textbf{0.5415} & \textbf{0.5412} & \textbf{0.4036} & \textbf{0.3257} & \textbf{0.4391} & \textbf{0.3906} \\
\textsc{\model} (Cross-Encoder, $P\leftarrow P\rightarrow P$)                                 & 0.7045          & 0.5356*         & 0.4168*         & 0.5944*         & 0.5336*         & 0.5243*         & 0.3932*         & 0.3190*         & 0.4271*         & 0.3814*         \\
\textsc{\model} (Cross-Encoder, $P(AA)P$)                                & 0.7028          & 0.5351          & 0.4171          & 0.5939          & 0.5338          & 0.5290*         & 0.3937          & 0.3201          & 0.4285*         & 0.3830          \\
\textsc{\model} (Cross-Encoder, $P(AV)P$)                                & 0.7024*         & 0.5354*         & 0.4177          & 0.5940*         & 0.5343*         & 0.5164**        & 0.3897*         & 0.3195*         & 0.4225*         & 0.3797*         \\
\textsc{\model} (Cross-Encoder, $P\rightarrow (PP)\leftarrow P$)                               & 0.7076*         & 0.5379*         & 0.4188          & 0.5971*         & 0.5363*         & 0.5186          & 0.3924*         & 0.3184*         & 0.4254*         & 0.3800*         \\
\textsc{\model} (Cross-Encoder, $P\leftarrow (PP)\rightarrow P$)                                & 0.7061          & 0.5376          & 0.4187          & 0.5964          & 0.5357          & 0.5218          & 0.3911          & 0.3172*         & 0.4249          & 0.3794          \\ \hline
\end{tabular}
}
\label{tab:path}
\end{table*}

Table \ref{tab:path} shows the performance of all 20 \textsc{\model} variants (2 architectures $\times$ 10 meta-paths/meta-graphs). We have the following observations: (1) All meta-paths and meta-graphs used in \textsc{\model}, except $PVP$, can improve the classification performance upon unfine-tuned SciBERT. For $PVP$, the unsatisfying performance is expected because venue information alone (e.g., \textit{ACL}) is too weak to distinguish between fine-grained labels (e.g., ``\texttt{Named Entity Recognition}'' and ``\texttt{Entity Linking}''). In Appendix \ref{sec:appen_inter}, we provide a mathematical interpretation of why some meta-paths/meta-graphs (e.g., $PVP$ or $P(AAAAA)P$) may not perform well within our \textsc{\model} framework. In short, the results in Table \ref{tab:path} demonstrate the effectiveness of \textsc{\model} across different meta-paths/meta-graphs. (2) Cross-Encoder models perform better than their Bi-Encoder counterparts in most cases (8 out of 10 meta-paths/meta-graphs on MAG-CS and 10 out of 10 on PubMed, in terms of P@1). 
(3) In contrast to the gap between Bi-Encoder and Cross-Encoder, the difference among citation-based meta-paths and meta-graphs is less significant. It would be an interesting future work to automatically select the most effective meta-paths/meta-graphs, although related studies on heterogeneous network representation learning \cite{dong2017metapath2vec,wang2019heterogeneous,yang2020heterogeneous} often require users to specify them.




\section{Related Work}
\noindent \textbf{Zero-Shot Multi-Label Text Classification.} Yin et al. \cite{yin2019benchmarking} divide existing studies on zero-shot text classification into two settings: the \textit{restrictive} setting \cite{wang2019survey} assumes training documents are given for some seen classes and the trained classifier should be able to predict unseen classes; the \textit{wild} setting does not assume any seen classes and the classifier needs to make prediction without any annotated training data. Most previous studies on zero-shot LMTC \cite{chalkidis2019large,nam2015predicting,nam2016all,rios2018few,gupta2021generalized} focus on the \textit{restrictive} setting, while this paper studies the more challenging \textit{wild} setting. Under the wild setting, a pioneering approach is dataless classification \cite{chang2008importance,song2014dataless} which maps documents and labels into the same space of Wikipedia concepts. Recent studies further leverage convolutional networks \cite{meng2018weakly} or pre-trained language models \cite{mekala2020contextualized,meng2020weakly,wang2020x}. While these models rely on label names only, they all assume that each document belongs to one category, thus are not applicable to the multi-label setting. Yin et al. \cite{yin2019benchmarking} propose to treat zero-shot text classification as a textual entailment problem and leverage pre-trained language models to solve it; Shen et al. \cite{shen2021taxoclass} further extend the idea to hierarchical zero-shot text classification. Compared with their studies, we utilize document metadata as complementary signals to the text.

\vspace{1mm}

\noindent \textbf{Metadata-Aware Text Classification.} In some specific classification tasks, document metadata have been used as label indicators (e.g., reviewer and product information in review sentiment analysis \cite{tang2015learning}, user profile information in tweet localization \cite{schulz2013multi,zhang2017rate}). Kim et al. \cite{kim2019categorical} propose a generic approach to add categorical metadata into neural text classifiers. Zhang et al. \cite{zhang2021match} further study metadata-aware LMTC. However, all these studies focus on the fully supervised setting. Some studies \cite{zhang2020minimally,zhang2021hierarchical,zhang2021minimally,zhang2021motifclass} leverage metadata in \textit{few-shot} text classification. Nevertheless, in LMTC, since the label space is large, it becomes prohibitive to provide even a few training samples for each label. Mekala et al. \cite{mekala2020meta} consider metadata in zero-shot single-label text classification given a small label space, but their approach can hardly be generalized to LMTC.

\vspace{1mm}

\noindent \textbf{Text Contrastive Learning.} Recently, contrastive learning has become a promising trend in unsupervised text representation learning. The general idea is to use various data augmentation techniques \cite{wei2019eda,xie2020unsupervised,feng2021survey} to generate similar text pairs and find a mapping function to make them closer in the representation space while pushing away dissimilar ones. Therefore, the major novelty of those studies is often the data augmentation techniques they propose. For example, DeCLUTR \cite{giorgi2021declutr} samples two text spans from the same document; CLEAR \cite{wu2020clear} adopts span deletion, span reordering, and synonym substitution; DECA \cite{luo2021unsupervised} uses synonym substitution, antonym augmentation, and back translation; SimCSE \cite{gao2021simcse} applies two different hidden dropout masks when encoding the same sentence; ConSERT \cite{yan2021consert} employs token reordering, deletion, dropout, and adversarial attack. However, all these data augmentation approaches manipulate \textit{text information only} to generate \textit{artificial} sentences/documents, while our \textsc{\model} exploits \textit{metadata} information to find \textit{real} documents that are similar to each other.

\section{Conclusions and Future Work}
In this paper, we study zero-shot multi-label text classification with document metadata as complementary signals, which avail us with heterogeneous network information besides corpus. Our setting does not require any annotated training documents with labels and only relies on label names and descriptions. We propose a novel metadata-induced contrastive learning (\textsc{\model}) method to train a BERT-based document--label relevance scorer. We study two types of architectures (i.e., Bi-Encoder and Cross-Encoder) and 10 different meta-paths/meta-graphs within the \textsc{\model} framework. \textsc{\model} achieves strong performance on two large datasets, outperforming competitive baselines under the zero-shot setting and being on par with the supervised MATCH model trained on 10K--200K labeled documents. In the future, it is of interest to extend \textsc{\model} to zero-shot hierarchical text classification and explore whether label hierarchy could provide additional signals to contrastive learning.


\newpage
\section*{Acknowledgments}
We thank anonymous reviewers for their valuable and insightful feedback.
Research was supported in part by US DARPA KAIROS Program No.\ FA8750-19-2-1004, SocialSim Program No.\ W911NF-17-C-0099, and INCAS Program No.\ HR001121C0165, National Science Foundation IIS-19-56151, IIS-17-41317, and IIS 17-04532, and the Molecule Maker Lab Institute: An AI Research Institutes program supported by NSF under Award No.\ 2019897. Any opinions, findings, and conclusions or recommendations expressed herein are those of the authors and do not necessarily represent the views, either expressed or implied, of DARPA or the U.S. Government.

\balance
\bibliographystyle{ACM-Reference-Format}
\bibliography{www}

\newpage
\nobalance
\appendix
\section{Supplementary Material}
\subsection{Interpretation of \textsc{\model}}
\label{sec:appen_inter}
We have proposed the \textsc{\model} objective (i.e., Eq. (\ref{eqn:bi}) or (\ref{eqn:cross})) to fine-tune BERT. Here, we provide the interpretation of this objective function by answering the following two questions.

\begin{itemize}[leftmargin=*]
    \item \textbf{Q1.} \textsc{\model} does not require any labeled data during fine-tuning. What is the relationship between \textsc{\model} and supervised fine-tuning approaches?
    \item \textbf{Q2.} Under what conditions will the \textsc{\model} objective be closer to the supervised objective (thus \textsc{\model} may perform better in classification)?
\end{itemize}

To answer these two questions, let us first consider how \textit{supervised} approaches fine-tune BERT in multi-label text classification. Assume each document $d \in \mathcal{D}$ is annotated with its relevant labels $\mathcal{L}_d$. We can sample a positive label $l^+$ from $\mathcal{L}_d$ and several negative labels $\{l_i^-\}_{i=1}^N$ from $\mathcal{L}$. The supervised learning process learns effective representations by pulling $d$ and $l^+$ together while pushing $d$ and $l_i^-$ apart. Using either Bi-Encoder (Eqs. (\ref{eqn:bifwd}) and (\ref{eqn:biscore})) or Cross-Encoder (Eqs. (\ref{eqn:crossfwd}) and (\ref{eqn:crossscore})), the learning objective can be defined as
\begin{equation}
\small
\mathcal{J}_{\rm Sup} = \mathop{\mathbb{E}}_{\substack{l^+ \in \mathcal{L}_d \\ l_i^-\sim \mathcal{L}}} \bigg[-\log\frac{\exp(score(d, l^+))}{\exp(score(d, l^+))+\sum_{i=1}^N \exp(score(d, l_i^-))}\bigg].
\label{eqn:sup1}
\end{equation}
If we view each label $l$'s text information (i.e., label name and description) as a ``document'', it is natural to assume this ``document'' $t_l$ is relevant to the label $l$. For example, in Figure \ref{fig:label}, the description of ``\texttt{Webgraph}'' can be viewed as a ``document'' specifically relevant to ``\texttt{Webgraph}'' itself. Therefore, if we follow the notation of $\mathcal{L}_d$ and use $\mathcal{L}_{l}$ to denote the set of labels relevant to $t_l$, it is equivalent to have $\mathcal{L}_{l}=\{l\}$. In this case, we have $l \in \mathcal{L}_d$ if and only if $\mathcal{L}_d \cap \mathcal{L}_{l} \neq \emptyset$. Then, the supervised loss in Eq. (\ref{eqn:sup1}) is equivalent to
\begin{equation}
\small
\mathcal{J}_{\rm Sup} = \mathop{\mathbb{E}}_{\substack{\mathcal{L}_d \cap \mathcal{L}_{l^+} \neq \emptyset \\ l_i^-\sim \mathcal{L}}} \bigg[-\log\frac{\exp(score(d, l^+))}{\exp(score(d, l^+))+\sum_{i=1}^N \exp(score(d, l_i^-))}\bigg].
\label{eqn:sup2}
\end{equation}
If we replace $l^+$ and $l_i^-$ in Eq. (\ref{eqn:sup2}) with $d^+$ and $d_i^-$, respectively, (in other words, if we replace label text information with real documents), the supervised loss can be rewritten into the following form.
\begin{equation}
\small
\mathcal{J}'_{\rm Sup} = \mathop{\mathbb{E}}_{\substack{\mathcal{L}_d \cap \mathcal{L}_{d^+} \neq \emptyset \\ d_i^-\sim \mathcal{D}}} \bigg[-\log\frac{\exp(score(d, d^+))}{\exp(score(d, d^+))+\sum_{i=1}^N \exp(score(d, d_i^-))}\bigg].
\label{eqn:sup3}
\end{equation}

By comparing Eq. (\ref{eqn:sup3}) with Eq. (\ref{eqn:bi}) (when $\tau=1$) and Eq. (\ref{eqn:cross}), we can answer \textbf{Q1}: \textsc{\model} and supervised loss share the same functional format, and the only difference is the distribution of training data $(d, d^+)$ pairs. For supervised loss, annotated documents are available, so positive document--document pairs $(d, d^+)$ can be derived using their labels; for \textsc{\model}, there are no annotated training samples, positive document--document pairs $(d, d^+)$ are derived from metadata instead.

Now we proceed to \textbf{Q2}. Given the answer of \textbf{Q1}, the key difference between $\mathcal{J}_{\rm Bi}$ (or $\mathcal{J}_{\rm Cross}$) and $\mathcal{J}'_{\rm Sup}$ is how we sample the positive partner $d^+$ for each document $d$. Let us explicitly write down the two distributions for sampling $d^+$. For \textsc{\model}, the distribution is
\begin{equation}
    {\rm P}_{\textsc{\model}}(d^+|d) = 
    \begin{cases}
    1/X, & d^+ \in \mathcal{N}_{\mathcal{M}}(d);\\
    0, & d^+ \notin \mathcal{N}_{\mathcal{M}}(d).
    \end{cases}
\end{equation}
Here, $X=|\mathcal{N}_{\mathcal{M}}(d)|$.

For the supervised loss in Eq. (\ref{eqn:sup3}), the distribution is
\begin{equation}
    {\rm P}_{\rm Sup}(d^+|d) = 
    \begin{cases}
    1/Y, & \mathcal{L}_d \cap \mathcal{L}_{d^+} \neq \emptyset;\\
    0, & \mathcal{L}_d \cap \mathcal{L}_{d^+} = \emptyset.
    \end{cases}
\end{equation}
Here, $Y=|\{d^+|\mathcal{L}_d \cap \mathcal{L}_{d^+} \neq \emptyset\}|$. 

It is straightforward to compute the Jensen-Shannon (JS) divergence between ${\rm P}_{\textsc{\model}}$ and ${\rm P}_{\rm Sup}$.
\begin{equation}
\begin{split}
    JS({\rm P}_{\textsc{\model}} || {\rm P}_{\rm Sup}) = & \frac{1}{2} \log\frac{2X}{X+Y} + \frac{1}{2} \sum_{\substack{d^+ \in \mathcal{N}_{\mathcal{M}}(d) \\ \mathcal{L}_d \cap \mathcal{L}_{d^+} = \emptyset}} \frac{1}{X} \log\Big(1+\frac{X}{Y}\Big) \ +  \\
    & \frac{1}{2} \log\frac{2Y}{X+Y} + \frac{1}{2} \sum_{\substack{\mathcal{L}_d \cap \mathcal{L}_{d^+} \neq \emptyset \\ d^+ \notin \mathcal{N}_{\mathcal{M}}(d)}} \frac{1}{Y} \log\Big(1+\frac{Y}{X}\Big).
\end{split}
\label{eqn:js}
\end{equation}
When $JS({\rm P}_{\textsc{\model}} || {\rm P}_{\rm Sup})$ is small, ${\rm P}_{\textsc{\model}}$ is close to ${\rm P}_{\rm Sup}$, thus the \textsc{\model} objective $\mathcal{J}_{\rm Bi}$ or $\mathcal{J}_{\rm Cross}$ is similar to the supervised objective $\mathcal{J}'_{\rm Sup}$.
To make $JS({\rm P}_{\textsc{\model}} || {\rm P}_{\rm Sup})$ smaller, the meta-path/meta-graph $\mathcal{M}$ should satisfy the following two properties as close as possible: 

\textbf{Property 1:} $d^+ \in \mathcal{N}_{\mathcal{M}}(d) \Rightarrow \mathcal{L}_d \cap \mathcal{L}_{d^+} \neq \emptyset$. Intuitively, this property means that if $d^+$ is reachable from $d$ via $\mathcal{M}$, then $d$ and $d^+$ should have similar topic labels. When this property is perfectly satisfied, the second term in Eq. (\ref{eqn:js}) is 0. 

\textbf{Property 2 (the inverse of Property 1):} $\mathcal{L}_d \cap \mathcal{L}_{d^+} \neq \emptyset \Rightarrow d^+ \in \mathcal{N}_{\mathcal{M}}(d)$. This property implies that if $d$ and $d^+$ have similar labels, then $d^+$ should be reachable from $d$ via $\mathcal{M}$. When this property is perfectly satisfied, the fourth term in Eq. (\ref{eqn:js}) is 0.

For example, when the label space is fine-grained, using $\mathcal{M}=PVP$ may not be a good choice because sharing the same venue is not sufficient to conclude that two papers have similar fine-grained labels. $PVP$ actually violates Property 1 in this case. On the other hand, when the label space is coarse-grained, using $\mathcal{M}=P(AAAAA)P$ may not be suitable because two papers in the same category do not necessarily have so many common authors. In this case, $P(AAAAA)P$ violates Property 2.

\subsection{Training and Inference Procedures of \textsc{\model}}
\label{sec:appen_sum}
\newlength{\textfloatsepsave} 
\setlength{\textfloatsepsave}{\textfloatsep}
\setlength{\textfloatsep}{0pt}
\begin{algorithm}[h]
\small
\caption{\textsc{\model} Training}
\label{alg:training}
\KwIn{
An unlabeled corpus $\mathcal{D}$ with metadata $G=(\mathcal{V}, \mathcal{E})$; \\
\ \ \ \ \ \ \ \ \ \ \ \ \ a meta-path/meta-graph $\mathcal{M}$; \\
\ \ \ \ \ \ \ \ \ \ \ \ \ the pre-trained BERT model.
}
\KwOut{A fine-tuned BERT model.}
Sample $d$, $d^+$, and $\{d_i^-\}_{i=1}^N$ from $\mathcal{D}$ according to $\mathcal{M}$\;
\If{using Bi-Encoder for fine-tuning}
{
    $\bme_d$, $\bme_{d^+}$, $\{\bme_{d_i^-}\}_{i=1}^N \gets$ Eq. (\ref{eqn:bien})\;
    Fine-tuning BERT by optimizing Eq. (\ref{eqn:bi})\;
}
\uElseIf{using Cross-Encoder for fine-tuning}
{
    $score(d, d^+)$, $score(d, d_i^-) \gets$ Eq. (\ref{eqn:crossen})\;
    Fine-tuning BERT by optimizing Eq. (\ref{eqn:cross})\;
}
Return the fine-tuned BERT\;
\end{algorithm}
\begin{algorithm}[h]
\small
\caption{\textsc{\model} Inference}
\label{alg:inference}
\KwIn{
A label space $\mathcal{L}$ with label names and descriptions $t_l$; \\
\ \ \ \ \ \ \ \ \ \ \ \ \ the fine-tuned BERT model; \\
\ \ \ \ \ \ \ \ \ \ \ \ \ a new document $d \notin \mathcal{D}$.
}
\KwOut{$d$'s relevant labels $\mathcal{L}_d \subseteq \mathcal{L}$.}
\textcolor{myblue}{// The Retrieval Stage\;}
$\mathcal{C}_{\rm exact}(d) \gets$ exact label name matching\;
$\mathcal{C}_{\rm BM25}(d) \gets$ Eq. (\ref{eqn:bm25})\;
$\mathcal{C}(d) = \mathcal{C}_{\rm exact}(d) \cup \mathcal{C}_{\rm BM25}(d)$\;
\textcolor{myblue}{// The Re-ranking Stage\;}
\For{$l \in \mathcal{C}(d)$}
{
\If{using Bi-Encoder}
{
    $\bme_d$, $\bme_l \gets$ Eq. (\ref{eqn:bifwd}) using the fine-tuned BERT model\;
    $score(d, l) \gets$ Eq. (\ref{eqn:biscore})\;
}
\uElseIf{using Cross-Encoder}
{
    $\bme_{d || t_l} \gets$ Eq. (\ref{eqn:crossfwd}) using the fine-tuned BERT model\;
    $score(d, l) \gets$ Eq. (\ref{eqn:crossscore})\;
}
}
Rank $l \in \mathcal{C}(d)$ according to $score(d, l)$\;
Return $\mathcal{L}_d =\{$top-$k$ ranked labels in $\mathcal{C}(d)\}$\;
\end{algorithm}
\setlength{\textfloatsep}{\textfloatsepsave}

We summarize the complete training and inference procedures in Algorithms \ref{alg:training} and \ref{alg:inference}, respectively. The goal of training is to obtain a fine-tuned BERT model, which is later used in the re-ranking stage of inference. Note that during the model training phase, we calculate the similarity score between two documents, while during the inference phase, we calculate the score between a document and a label (represented by its name and description).

\subsection{Implementation}
\label{sec:appen_impl}
\subsubsection{Optimization of \textsc{\model}} 
To approximate the expectation in Eqs. (\ref{eqn:bi}) and (\ref{eqn:cross}) during optimization, we adopt a sampling strategy. Specifically, we first sample a set of documents from $\mathcal{D}$. For each sampled document $d$, we pick one $d^+$ from $\mathcal{N}_{\mathcal{M}}(d)$ as $d$'s positive partner (if $\mathcal{N}_{\mathcal{M}}(d) \neq \emptyset$). When fine-tuning the \textit{Bi-Encoder}, following \cite{chen2020simple}, $d$'s negative partners $\{d_i^-\}_{i=1}^N$ are the positive partner of other documents in the same training batch. Each $d$ has 1 positive partner and $(\beta-1)$ negative partners, where $\beta$ is the batch size. Therefore, within each batch, to calculate Eq. (\ref{eqn:bi}), we need to call BERT $2\beta$ times (i.e., computing $\bme_d$ and $\bme_{d^+}$ for each $d$). When fine-tuning the \textit{Cross-Encoder}, however, we would call BERT $\beta^2$ times within each batch if we used the same approach to obtain negative partners. To make the optimization process more efficient, for each $d$, we directly sample one negative partner $d^-$ from $\mathcal{D}$. Each $d$ has 1 positive partner and 1 negative partner. In this way, we only need to call BERT $2\beta$ times within each batch (i.e., encoding $(d\ ||\ d^+)$ and $(d\ ||\ d^-)$ for each $d$).

\subsubsection{Parameter Settings of \textsc{\model}}
For both Bi-Encoder and Cross-Encoder, we sample 50,000 $(d, d^+)$ pairs for training and 5,000 $(d, d^+)$ pairs for validation. The training batch size $\beta$ is $8$ and $4$ for Bi-Encoder and Cross-Encoder, respectively. The maximum length of SciBERT in Bi-Encoder is 256; the maximum length of SciBERT in Cross-Encoder is 512 (i.e., 256 for each document before concatenation). The temperature hyperparameter $\tau=0.05$. We train the model for 3 epochs using Adam as the optimizer. During inference, the threshold of BM25 scores is $\eta=400$.

\subsection{Datasets}
\label{sec:appen_data}
Table \ref{tab:data} mainly summarizes text and label statistics (where Labels/Doc and Words/Doc refer to the statistics in the testing set). In Table \ref{tab:appen_data}, we list metadata-related statistics of the training corpus $\mathcal{D}$.

\begin{table}[!h]
\small
\centering
\caption{Metadata-related statistics of the training corpus.}
\vspace{-0.5em}
\scalebox{0.88}{
\begin{tabular}{c|cc}
\hline
\textbf{Dataset}               & \textbf{MAG-CS \cite{wang2020microsoft}}   & \textbf{PubMed \cite{lu2011pubmed}}   \\ \hline
\# Authors            & 762,259   & 2,068,411 \\
\# Author--Paper Edges & 2,047,166 & 5,391,314 \\
\# Venues             & 105       & 150       \\
\# Venue--Paper Edges  & 634,874   & 808,692   \\
\# Paper$\rightarrow$Paper Edges  & 1,219,234 & 3,615,220 \\ \hline
\end{tabular}
}
\label{tab:appen_data}
\vspace{-0.5em}
\end{table}

\subsection{Additional Experiments: Combining Top-Performing Meta-Paths/Meta-Graphs}
\begin{figure}[!h]
\centering
\subfigure[MAG-CS, Cross-Encoder]{
\includegraphics[width=0.23\textwidth]{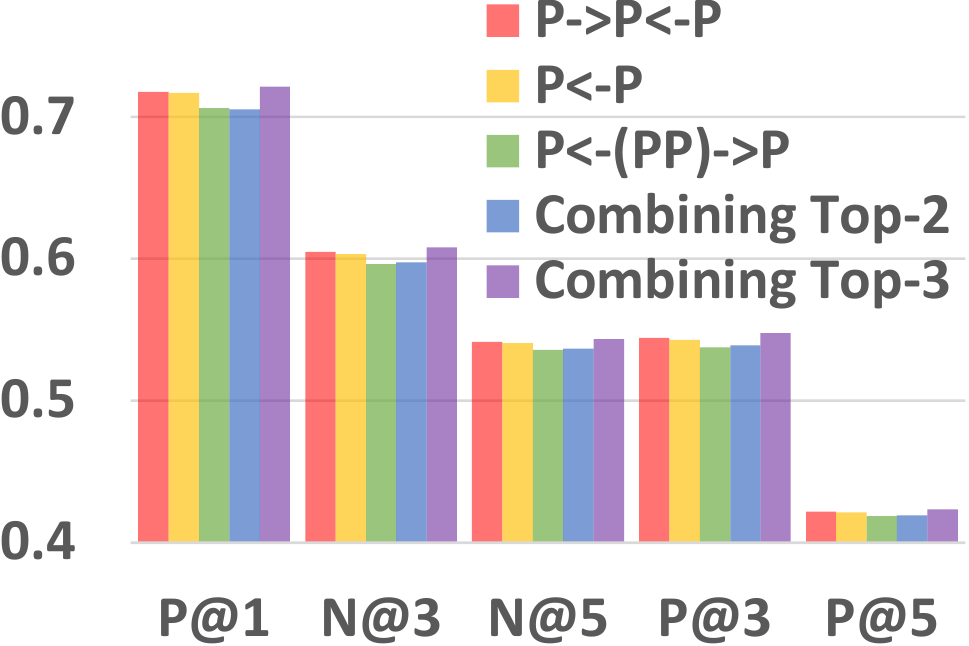}}
\hspace{-1mm}
\subfigure[PubMed, Cross-Encoder]{
\includegraphics[width=0.23\textwidth]{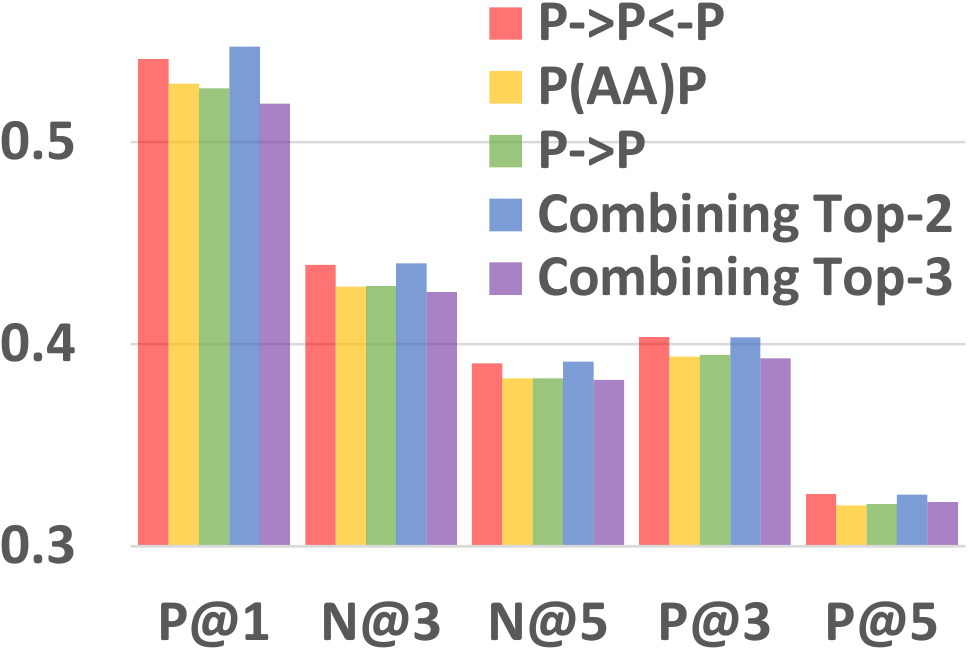}}
\vspace{-1em}
\caption{P@$k$ and NDCG@$k$ scores of top-3 performing meta-paths/meta-graphs and the \textit{combination} of them.} 
\vspace{-1em}
\label{fig:merge}
\end{figure}

Figure \ref{fig:merge} compares P@$k$ and NDCG@$k$ scores of top-3 performing meta-paths/meta-graphs and the \textit{combination} of them. We \textit{combine} them by putting document--document pairs induced by different meta-paths/meta-graphs together to train the model. In Figure \ref{fig:merge}, we cannot observe consistent and significant benefit of \textit{combining} meta-paths/meta-graphs, possibly because two documents judged as similar by one meta-path may be viewed as dissimilar by another, which may confuse our model during training.

\end{spacing}
\end{document}